\documentclass[10pt,twocolumn,letterpaper]{article}
\usepackage[accsupp]{axessibility}
\usepackage{iccv}
\usepackage{times}
\usepackage{epsfig}
\usepackage{graphicx}
\usepackage{amsmath}
\usepackage{amssymb}

\usepackage{colortbl}

\usepackage{graphicx}
\usepackage{array}
\usepackage{adjustbox}
\usepackage{multirow}
\usepackage{colortbl}

\usepackage{subcaption}
\usepackage[size=small]{caption}
\usepackage{color,xcolor}
\usepackage{epsfig}
\usepackage{booktabs}
\usepackage[square,sort,comma,numbers]{natbib}

\usepackage{hhline}

\usepackage{amsmath,amsfonts,amsthm,amssymb}
\usepackage{mathtools}
\usepackage{bm}
\usepackage{nicefrac}
\usepackage{microtype}

\usepackage[pagebackref=true,breaklinks=true,colorlinks,citecolor=gray]{hyperref}
\usepackage{algorithm, algorithmic}
\newcolumntype{L}[1]{>{\raggedright\let\newline\\\arraybackslash\hspace{0pt}}m{#1}}
\newcolumntype{C}[1]{>{\centering\let\newline\\\arraybackslash\hspace{0pt}}m{#1}}
\newcolumntype{R}[1]{>{\raggedleft\let\newline\\\arraybackslash\hspace{0pt}}m{#1}}

\newcommand{\sect}[1]{Section~\ref{#1}}

\newcommand{\eqn}[1]{Equation~\ref{#1}}
\newcommand{\fig}[1]{Figure~\ref{#1}}
\newcommand{\tbl}[1]{Table~\ref{#1}}
\newcommand{\alg}[1]{Algorithm~\ref{#1}}

\newcommand{\degree}{\ensuremath{^\circ}\xspace}
\newcommand{\ignore}[1]{}

\makeatletter
\DeclareRobustCommand\onedot{\futurelet\@let@token\@onedot}
\def\@onedot{\ifx\@let@token.\else.\null\fi\xspace}

\def\etal{\emph{et al}\onedot}
\makeatother

\definecolor{MyDarkBlue}{rgb}{0,0.08,1}
\definecolor{MyDarkGreen}{rgb}{0.02,0.6,0.02}
\definecolor{MyDarkRed}{rgb}{0.8,0.02,0.02}
\definecolor{MyDarkOrange}{rgb}{0.40,0.2,0.02}
\definecolor{MyPurple}{RGB}{111,0,255}
\definecolor{MyRed}{rgb}{1.0,0.0,0.0}
\definecolor{MyGold}{rgb}{0.75,0.6,0.12}
\definecolor{MyDarkgray}{rgb}{0.66, 0.66, 0.66}

\newcommand{\myparagraph}[1]{\vspace{0.1em} \noindent \textbf{#1}}

\newcommand{\model}{CRL\xspace}

\DeclarePairedDelimiterX{\infdivx}[2]{(}{)}{%
  #1\;\delimsize|\delimsize|\;#2%
}

\usepackage{amsmath,amsfonts,bm}

\def\eqref#1{equation~\ref{#1}}
\def\1{\bm{1}}

\newcommand{\test}{\mathcal{D_{\mathrm{test}}}}

\def\vx{{\bm{x}}}

\def\vz{{\bm{z}}}

\DeclareMathAlphabet{\mathsfit}{\encodingdefault}{\sfdefault}{m}{sl}
\SetMathAlphabet{\mathsfit}{bold}{\encodingdefault}{\sfdefault}{bx}{n}

\usepackage{comment}
\usepackage{subcaption}
\iccvfinalcopy

\def\iccvPaperID{3922} % *** Enter the ICCV Paper ID here

\ificcvfinal\fi

\begin{document}
% \linenumbers

%%%%%%%%% TITLE
% \title{Curious Representation Learning}
\title{Curious Representation Learning for Embodied Intelligence}
% \title{Appendix: Curious Representation Learning for Embodied Intelligence}
%\title{Curious Representation Learning: Training Representations that Transfer to Interactive Tasks}
%\title{Curious Representation Learning in Embodied Environments}
%\title{Curious Representation Learning for Embodied Intelligence}

\author{Yilun Du\\
MIT\\
\and
Chuang Gan\\
MIT-IBM Watson AI Lab\\
\and
Phillip Isola\\
MIT\\
}

\maketitle

% \pagestyle{headings}
% \mainmatter
% \def\ECCVSubNumber{7496}  % Insert your submission number here

% \title{Visual Representation Learning in Embodied Environments} % Replace with your title
% %\title{Visual Representation Learning in Embodied Environments}
% %\title{Learning Visual Representations from Environments rather than from Datasets}
% %\title{Curious Representation Learning}

% % INITIAL SUBMISSION 
% %\begin{comment}
% \titlerunning{ECCV-20 submission ID \ECCVSubNumber} 
% \authorrunning{ECCV-20 submission ID \ECCVSubNumber} 
% \author{Anonymous ECCV submission}
% \institute{Paper ID \ECCVSubNumber}
% %\end{comment}
% %******************

% % CAMERA READY SUBMISSION
% \begin{comment}
% % If the paper title is too long for the running head, you can set
% % an abbreviated paper title here
% \end{comment}
% %******************
% \maketitle

\begin{abstract}
    Self-supervised representation learning has achieved remarkable success in recent years. By subverting the need for supervised labels, such approaches are able to utilize the numerous unlabeled images that exist on the Internet and in photographic datasets. Yet to build truly intelligent agents, we must construct representation learning algorithms that can learn not only from \textbf{datasets} but also learn from  \textbf{environments}. An agent in a natural environment will not typically be fed curated data. Instead, it must explore its environment to acquire the data it will learn from. We propose a framework, curious representation learning (CRL), which jointly learns a reinforcement learning policy and a visual representation model. The policy is trained to maximize the error of the representation learner, and in doing so is incentivized to explore its environment. At the same time, the learned representation becomes stronger and stronger as the policy feeds it ever harder data to learn from. Our learned representations enable promising transfer to downstream navigation tasks, performing better than or comparably to ImageNet pretraining without using any supervision at all. In addition, despite being trained in simulation, our learned representations can obtain interpretable results on real images.  Code is available at \url{https://yilundu.github.io/crl/}.
    % %n addition, the learned policies show effective exploration of their environments, and achieve competitive performance on benchmark tasks. %to actively explore the world with good visual representations.
\end{abstract}

\section{Introduction}

% Self-supervised representation learning has been highly successful in both natural language processing \citep{devlin2018bert, radford2019language} and computer vision \citep{he2019momentum}. 

Similar to biological agents, self-supervised agents learn representations without explicit supervisory labels \citep{smith2005development}. Impressively, these methods can surpass those based on supervised learning \citep{devlin2018bert}. Yet the most successful approaches also depart from biological learning in that they depend on a curated dataset of observations to learn from.

 \begin{figure}[t]
\begin{center}
\includegraphics[width=1.0\linewidth]{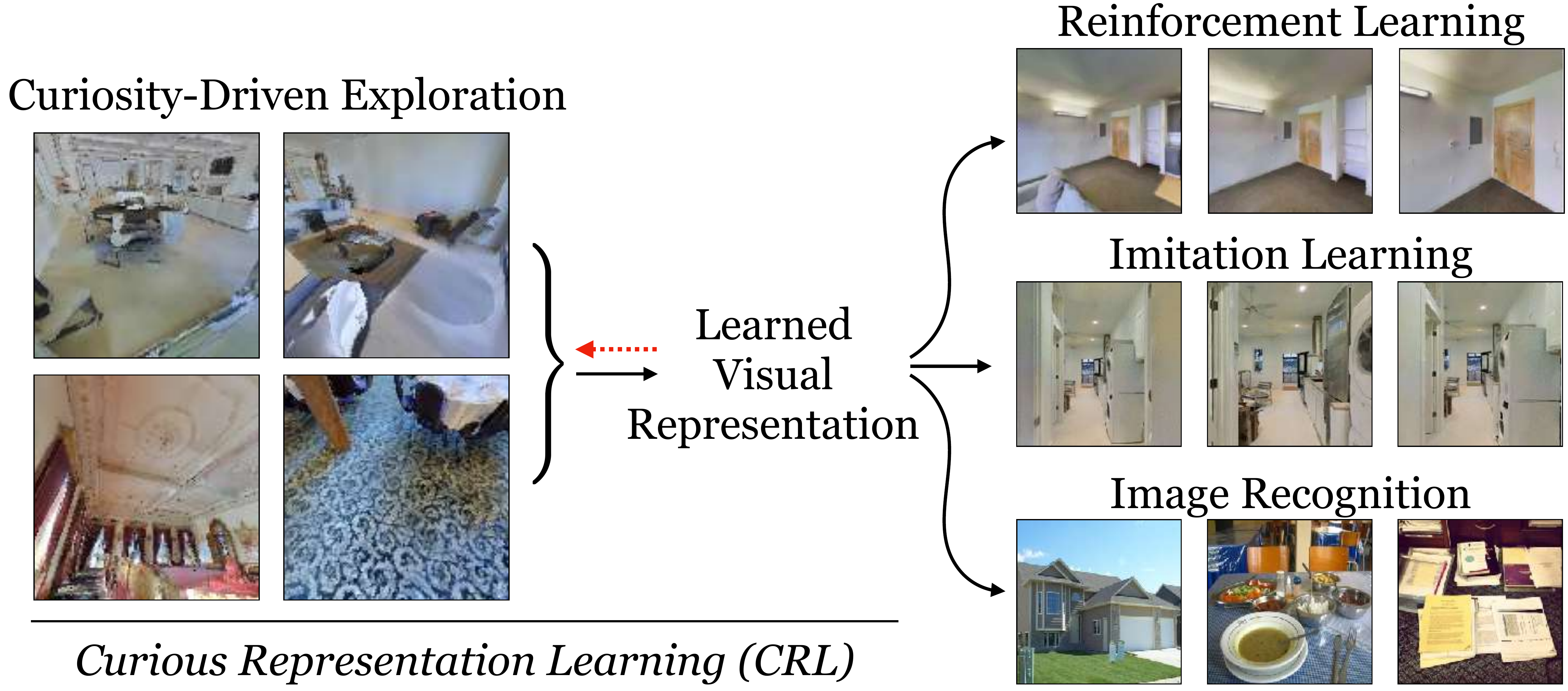}
\end{center}
\vspace{-7pt}

\caption{\small When put in a new world, without an explicit task or goal, we are still able to actively explore and interact with our surroundings. Our framework, \model, enables agents to learn visual representations from interaction without any supervision, using only curiosity-driven exploration where the agent seeks out observations that incur high error under the representation model. The resultant representations enable agents to perform well in downstream reinforcement and imitation learning tasks, and further are able to transfer to recognition of real images.}
\label{fig:teaser}
\vspace{-12pt}
\end{figure}

In stark contrast, learning in biological vision involves active physical exploration of an environment. Infants are not endowed with existing visual experience, but must instead explore to obtain such experience from the surrounding environment. By playing with toys, through actions such as pushing, grasping, sucking, or prodding, infants are able to obtain experiences of texture, material, and physics~\cite{gopnik1999scientist}. By crawling into new rooms, toddlers obtain experiences of layout and geometry. This setup adds additional challenges towards learning visual representations.  Algorithms must now selectively explore and determine which portion of the environment will allow for the most useful increase in visual experience. Furthermore, algorithms must also adapt to constant domain shift; at any time point, the only observed visual experience is that of a particular room, or that of a particular object being interacted with.

Given an interactive environment, and no prior data or tasks, how may we obtain a good visual representations? This is a challenging question and requires an agent to answer several subquestions. In particular, how can we learn to effectively explore and perceive the surrounding world? And, how can we integrate each different experience together to obtain the best representation possible? In this paper, we propose a unified framework towards solving these tasks.

One approach is to train a vision-based reinforcement learning agent in the interactive environment. Intuitively, as the agent learns to interact in its surrounding environment, its underlying vision system must also learn to understand the surrounding environment. A core difficulty, however, is the noisy and sparse supervision that reinforcement learning provides, inhibiting the formation of a strong vision system. An alternative approach is thus to leverage self-supervised representation learning techniques to learn representations in embodied environments. To gather data to learn the representation, a separate exploration algorithm may be used. However, such an approach raises additional challenges. Given a new embodied environment, how can we learn to effectively explore and obtain diverse data to train our representation?  And how can we continuously gather images that remain visually salient to our algorithm?

To address these issues, we propose a unified framework,  Curious Representation Learning (\model \fig{fig:teaser}). Our key idea is to automatically learn an exploration policy given a self-supervised representation learning technique by training a reinforcement learning (RL) to maximize reward equal to the loss of the self-supervised representation learning model. We then train our self-supervised model by minimizing the loss on the images obtained by the exploration policy. By defining the reward to our exploration policy in such a manner, it serves as a natural measure of visual novelty, as only on unfamiliar images will the loss be large. Thus, our policy learns to explore the surrounding environment and obtain images that are visually distinct from those seen in the past. Simultaneously, our self-supervised model benefits from diverse images, specifically obtained to remain visually salient to the model. 

Given an embodied visual representation, we further study how it may be used for downstream interactive tasks.  Interactive learning, through either reinforcement learning or behavioral cloning, is characterized by both sparse and noisy feedback. Feedback from individual actions is delayed across time and dependent on task completion, with feedback containing little information in the case a task failure, and giving conflicting results when other actions effect task completion. Such noise can quickly destroy learned visual representations. We find to enable good downstream interactive transfer, it is crucial to freeze visual network weights before transfer. We observe that our method can significantly boost the semantic navigation performance of RL policies and visual language navigation using imitation learning.

Our contributions in this paper are three-fold. First, we introduce \model as an approach to embodied representation learning, in which a representation learning model plays a minimax game with an exploration policy. Second, we show that learned visual representations can help in a variety of embodied tasks, where it is crucial to freeze representations to enable good performance. Finally, we show that our representations, while entirely trained in simulation, can obtain interpretable results on real photographs.

\section{Related Work}

\myparagraph{Self-supervised Visual Representation Learning:} Unsupervised representation learning has seen increased interest in recent years \citep{Bengio2013Representation,Zhang2016Colorful, gidaris2018unsupervised, hjelm2018learning}. Approaches towards unsupervised learning include colorizing images \citep{Zhang2016Colorful}, predicting image rotations \citep{gidaris2018unsupervised} and geometry transformation~\cite{gan2018geometry}, solving jigsaw puzzles \citep{noroozi2016unsupervised}, and adversarial inference \citep{dumoulin2016adversarially}. Recently, approaches based on maximizing mutual information have achieved success \citep{hjelm2018learning, oord2018representation, bachman2019learning, tian2019contrastive, henaff2019data, he2019momentum, chen2020simple}. While previous approaches have considered learning visual representations in static datasets, we consider the distinct problem of obtaining visual representations in interactive environments, where an agent must actively obtain data it is trained on. We provide a framework for learning a task-agnostic representation for different downstream interactive tasks.

\myparagraph{Curiosity-based Learning:} Our approach is also related to existing work in curiosity. Curiosity has also been studied extensively in the past years \citep{schmidhuber1990making,schmidhuber1991possibility, houthooft2016vime, Pathak2017Curiosity, bellemare2016unifying, burda2018exploration,gan2020noisy}, as both an incentive for exploration as well as a means of achieving emergent complex behavior. Recent works have formulated curiosity as a reward dependent on a learned model, such as an inverse dynamics model \citep{Pathak2017Curiosity}, learned features in a VAE \citep{burda2018large}, and features from a random network \citep{burda2018exploration}. In contrast to these work, which often rely on heuristic design choices to select rewards for each task, we construct curiosity as a minimax game between a generic representation learning algorithm and a reinforcement learning policy. This formulation then allows us to substitute existing representation learning algorithms into a curiosity-based formulation, enabling us to combine advances in representation learning and curiosity-based exploration.

\myparagraph{Embodied Representation Learning:}
Pinto et al. \cite{Pinto2016Curious} investigated representations that emerge from physical interaction in robotics. In contrast to our work, interactions are manually designed. Agrawal et al. \citep{Agrawal2016Learning} investigated emergent physical representations from poking in robotics. However, interactions are randomly generated and limited only to poking. Overall, our work focuses on both learning to interact and on representations in an interactive environment, and further focuses on reusing them for downstream applications. 

\myparagraph{Representation learning in RL:} Recently, using unsupervised/self-supervised representation learning methods to improve sample efficiency and/or performance in RL has gained increased popularity~\cite{jaderberg2016reinforcement, anand2020unsupervised, srinivas2020curl, stooke2020decoupling, schwarzer2020dataefficient}.  In contrast to prior work, which focuses on synthetic game environments, we study representation learning in photorealistic 3D environments~\cite{habitat19arxiv,xia2018gibson,gan2020threedworld}. 

Perhaps most related to our work is the work of Ye~\cite{ye2020auxiliary}, who show that using auxiliary tasks can improve PointGoal navigation results in the Gibson environment~\cite{xiazamirhe2018gibsonenv}. Different from them, we mainly investigate if we can, in a self-supervised manner, learn a generic and task-agnostic representation that can be reused for downstream interactive tasks. Concurrent to our work, Ramakrishnan \etal \citep{ramakrishnan2021environment} also investigate learning environment-level representations through environment predictive coding and show benefits to downstream visual exploration tasks. However, it remains unclear whether these learned representations could help in more challenging navigation tasks.    

% \myparagraph{Active Learning:} Finally, our work is related to past work in active learning. In active learning, datapoints  are sampled so that they maximize an uncertainty/error measure, while the learner itself thereby aims to minimize un-certainty/error \citep{settles2009active, roy2001toward}. This entails an adversarial game where as learners get better and better, it becomes increasingly difficult to train on, an idea refer ed to as curriculum learning \citep{bengio2009curriculum}. In our work, we explicitly formulate this arms race as a minimax optimization problem with a visual representation learning model.

\section{Curious Representation Learning}
\vspace{-5pt}
\begin{figure}[t]
\begin{center}
\includegraphics[width=1.0\linewidth]{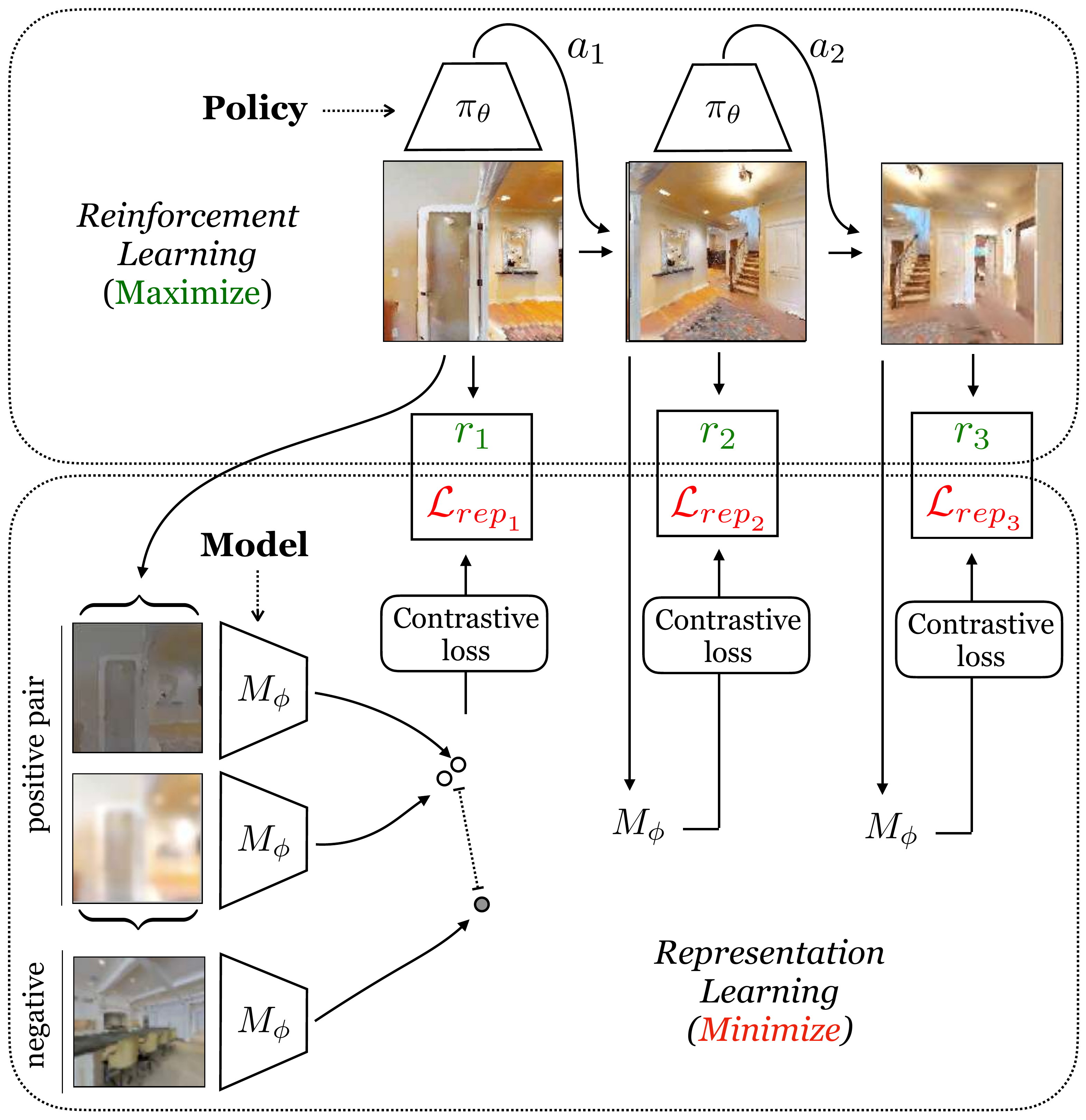}
\end{center}
\vspace{-15pt}
\caption{\small Overview of CRL (curious representation learning).  We jointly train a RL policy and visual representation learning model to learn visual representations in interactive environments. The RL policy and visual representation learning model engage in a mini-max game where the policy maximizes reward which is set to representation learning model's loss, while the model minimizes its own loss. For the representation learning model we use the contrastive learning method SimCLR~\cite{chen2020simple}. In the figure, we only diagram the full contrastive setup for the first frame, but note that it is applied to each frame.}
\label{fig:main_overview}
\vspace{-15pt}
\end{figure}

Our goal is to study how to obtain a generic, task-invariant representation for downstream interactive learning in an embodied environment,  without either task or extrinsic reward specification. We propose curious representation learning (\model), a unified framework for learning visual representations. We first review some background knowledge on the contrastive representation learning framework, and next describe \model, which extends any generic representation learning  objective to interactive environments. We then describe our overall policy and model optimization procedure and evaluation protocol and provide psuedocode in the appendix.

\subsection{Contrastive Representation Learning}

\label{sect:rep}
To learn representations, we utilize contrastive learning \citep{oord2018representation, he2019momentum, chen2020simple, bachman2019learning, tian2019contrastive, wu2018unsupervised}. 
Following \citep{chen2020simple}, our contrastive learning setup consists of a representation learning model $M_\phi$ and 2 layer MLP projection head $g_\psi$, and a family of data augmentations $\mathcal{T}$.

For a set of $N$ images, $\{\vx_k\}_{k=1\ldots N}$, we sample $2N$ different augmentations from $\mathcal{T}$, and apply two separate augmentations to each image to obtain pairs of augmented images $\{\tilde{\vx}_k^1, \tilde{\vx}_k^2\}_{k=1\ldots N}$. For a given image $\vx$, we obtain a latent representation $z$ by applying $\vz = \text{Normalize}(g_\psi(M_\phi(\vx)))$, where we L2 normalize the output of the projection head.

We then train our contrastive loss utilizing the InfoNCE loss \citep{oord2018representation}, which consists of

\begin{equation}
    \mathcal{L}_{\text{contrast}} = - \frac{1}{N} \sum_{i=1}^N \log \frac{\exp(\text{sim}(\tilde{\vz}_i^1, \tilde{\vz}_i^2) / \tau)}{\Sigma_{j,k=1}^N \exp(\text{sim}(\tilde{\vz}_j^1, \tilde{\vz}_k^2) / \tau)}
    \label{eqn:contrast}
\end{equation}

where $\text{sim}(\vz_i, \vz_j)$ corresponds to the dot product between latents $\vz_i$ and $\vz_j$. We utilize  $\tau=0.07$ and define $\mathcal{T}$ to consist of horizontal flips, random resized crops, and color saturation, using default parameters from \citep{chen2020simple}.

\subsection{Learning Representations from Intrinsic Motivation}

When representation learning is applied to a static dataset, a model $M_\phi$ is trained to minimize a representation learning $\mathcal{L}_{\text{rep}}$ objective (with the objective \eqn{eqn:contrast} corresponding to contrastive representation learning) over observed images $\vx$,  where images are drawn from a data distribution $p_{\text{data}}$%
\begin{align}
\min_{\phi} \underset{\vx \sim p_{\text{data}}}{\mathbb{E}}\left [\mathcal{L}_{\text{rep}}(M_\phi, \vx) \right] \label{eq:vrl}.
\end{align}
In contrast, in our interactive setting, images in our data distribution must now be actively chosen by an agent. We utilize a reinforcement learning policy $\pi_\theta$ to represent our agent, where a policy is trained to maximize  reward $r_t$ at each timestep $t$%
\begin{align}
\max_{\theta} \underset{\vx \sim \pi_\theta}{\mathbb{E}}\left [\sum_{t=0}^T r_t \right] \label{eq:rl}.
\end{align}
In our setting, we do not have access to an underlying task or reward, so we need to define implicitly define our reward. In \model, we note that we can directly use \eqn{eq:vrl} to define the reward function at each time step to train our reinforcement learning policy. Specifically, we use the loss of the representation learning criterion to be our reward so that our reinforcement learning objective is now:%
\begin{align}
\max_{\theta} \underset{\vx \sim \pi_\theta}{\mathbb{E}}\left [\sum_{t=0}^T \mathcal{L}_{\text{rep}}(M_\phi, \vx) \right] \label{eq:crl} .
\end{align}
This objective then encourages our policy to find images that $M_\phi$ incurs high losses on, giving a natural incentive for our policy to obtain interesting data to train our representation learning model.

Furthermore, note that while \eqn{eq:vrl} minimizes $\mathcal{L}_{\text{rep}}$ using $M_\phi$, \eqn{eq:crl} maximizes $\mathcal{L}_{\text{rep}}$ using $\pi_\theta$, leading to an overall mini-max game objective%
\begin{align}
\max_{\theta} \min_{\phi} \underset{\vx \sim \pi_\theta}{\mathbb{E}} \left[\sum_{t=0}^{T} \mathcal{L}_{\text{rep}}(M_\phi, \vx) \right].
\end{align}%
This mini-max game can be seen as a synergistic way to improve both a policy and a representation learning model. This new objective encourages our policy $\pi_\theta$ to learn complex navigation navigation and perception patterns so that it can effectively obtain images to surprise the representation learning model $M_\phi$. Simultaneously, this also allows our representation learning model to learn good representations that are resistant to samples found from the policy $\pi_\theta$. 

Our formulation of \model is similar to prior work in intrinsic motivation and curiosity \citep{Pathak2017Curiosity}. Such papers encourage reinforcement learning agents to explore  by giving agents reward equal some predictive loss.   By interpreting the representation learning loss as a predictive loss,  \model can thus be seen a curiosity model. However, while different past papers have proposed separate objectives for predictive error, such as  random features \citep{burda2018exploration} and inverse dynamics \citep{Pathak2017Curiosity}, \model provides a generic framework to further construct different curiosity objectives, by utilizing different representation learning models, automating the traditionally hand-designed process. Furthermore, \model allows us to reinterpret existing curiosity objectives as different methods to obtain an underlying representation of the world.

\subsection{Model and Policy Optimization}

When training our policy, we found that directly defining our reward following \eqn{eqn:contrast} led to a failure case where $\mathcal{L}_{\text{contrast}}$ loss could be maximized by having an agent stand in space (as all identical image observations maximizes the denominator of $\mathcal{L}_{\text{contrast}}$). To remedy this issue, we define the reward to our policy as only be the numerator of $\mathcal{L}_{\text{contrast}}$, $r_t = -\text{sim}(\vx^1, \vx^2)$. We further add a constant of 1 to all rewards to ensure that rewards at each observed image is non-negative. Furthermore, following \citep{burda2018large}, we normalize rewards by the standard deviation of past observed rewards to ensure that reward magnitudes do not change significantly.

Given computed rewards $r^{\gamma}_t$,  we use the proximal policy optimization (PPO) \citep{Schulman2017Proximal},  to train our policy and optimize the objective
$ L(\theta) = \mathbb{E} [\min (c_t(\theta) A_t, \text{clip} (c_t(\theta), 1-\epsilon, 1+\epsilon)A_t]$, where the clip ratio $c_t = \tfrac{\pi_{\theta}(a_t|s_t)}{\pi_{\theta_{\text{old}}}(a_t|s_t)}$ and  the advantage, $A_t$, is computed using the value function $V(s_t)$. We optimize both $\pi_\theta$ and $M_\phi$ using collected minibatches of data from PPO. Please see the appendix for pseudocode.

\subsection{Experimental Protocol}

Here we describe the protocols used for our empirical experiments. First, we discuss our protocol for learning embodied representations, and then our protocol for validating the utility of learned embodied representations to downstream tasks. 

\myparagraph{Representation Pretraining.} To pretrain representations, we train \model on Habitat simulator using the Matterport3D dataset \citep{chang2017matterport3d} for 10 million interactions, reserving the Gibson dataset \citep{xiazamirhe2018gibsonenv} for experimental validation. We train agents with 16 environments in parallel. Our observation space consists of only $256 \times 256$ RGB observations, and our action space consists of actions move forward by 0.25 meters, turn left by $30\degree$, turn right by $30\degree$, look up by $10\degree$ and look down by $10\degree$, with a maximum episode length of 500 steps.

\myparagraph{Downstream Evaluation.} We evaluate pretrained representations on the downstream tasks of semantic navigation, visual language navigation, and real image understanding.
For semantic navigation, we train a reinforcement learning agent using the Habitat simulator on the Gibson dataset  and on object navigation using the Habitat Matterport3D dataset (due to the lack of object annotations in Gibson).  For visual language navigation, we utilize an imitation learning agent on Matterport3D dataset, and on real images we utilize the Places dataset. We use the default environment settings for both in Habitat \citep{habitat19arxiv}. We utilize the \textit{same} representation across different tasks using the features from the last final average pooling layer of a ResNet50.  To enable effective interactive downstream transfer, we found that it was \textit{crucial} to freeze visual representations, due to the noisy nature of gradients from interactive tasks. Such a technique has also been noted to be useful in few-shot learning \citep{tian2020rethinking}.

\myparagraph{Model Architectures.} For representation learning models $M_\phi$ and policies $\pi_\theta$, we utilize a ResNet50 \citep{He2015Deep} image encoder. To enable stable reinforcement learning, we replace batch normalization layers with group normalization. To train $M_\phi$, we use a 2 layer projection head, with a projection dimension of 128 dimensions.

\section{Experiments}

We quantitatively and qualitatively show that \model can learn generic, task-agnostic visual representations for downstream interactive tasks. We discuss our experimental setup in \sect{sect:embodied_rep}. Next we analyze the interactive behavior of \model in \sect{sect:visual_explore}. Using one unified pretrained model, we show that we can improve performance in semantic navigation in RL in \sect{sect:rl}, visual-language navigation using imitation learning in \sect{sect:imitation}, and can further achieve transfer to recognition of real images in \sect{sect:real_image}. Finally, we discuss how we may also obtain representations in a real biological setting in the appendix.

\begin{figure}
\begin{center}
\includegraphics[width=0.8\linewidth]{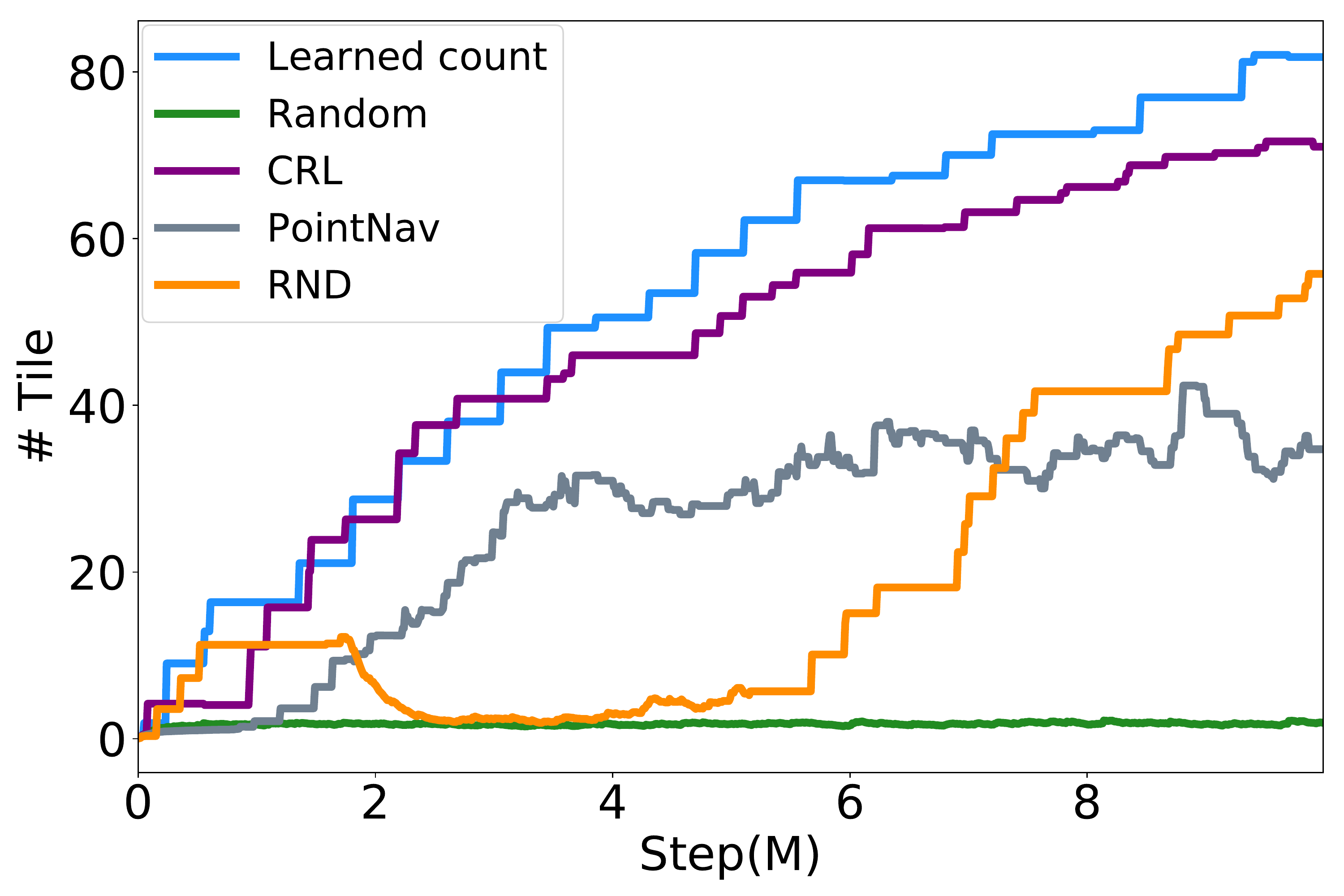}
\end{center}
\vspace{-20pt}
\caption{\small Plots of the average number of tiles explored in a across separate environments of different reinforcement learning agents. To gather images that have high contrastive loss, \model explores effectively around the surrounding environment, outperforming RND and PointNav agents, and performs similar to a learned counts agent that is  explicitly encouraged to maximize tile exploration.}
\label{fig:num_tile}
\vspace{-5pt}
\end{figure}
\begin{figure}
\begin{center}
\includegraphics[width=0.8\linewidth]{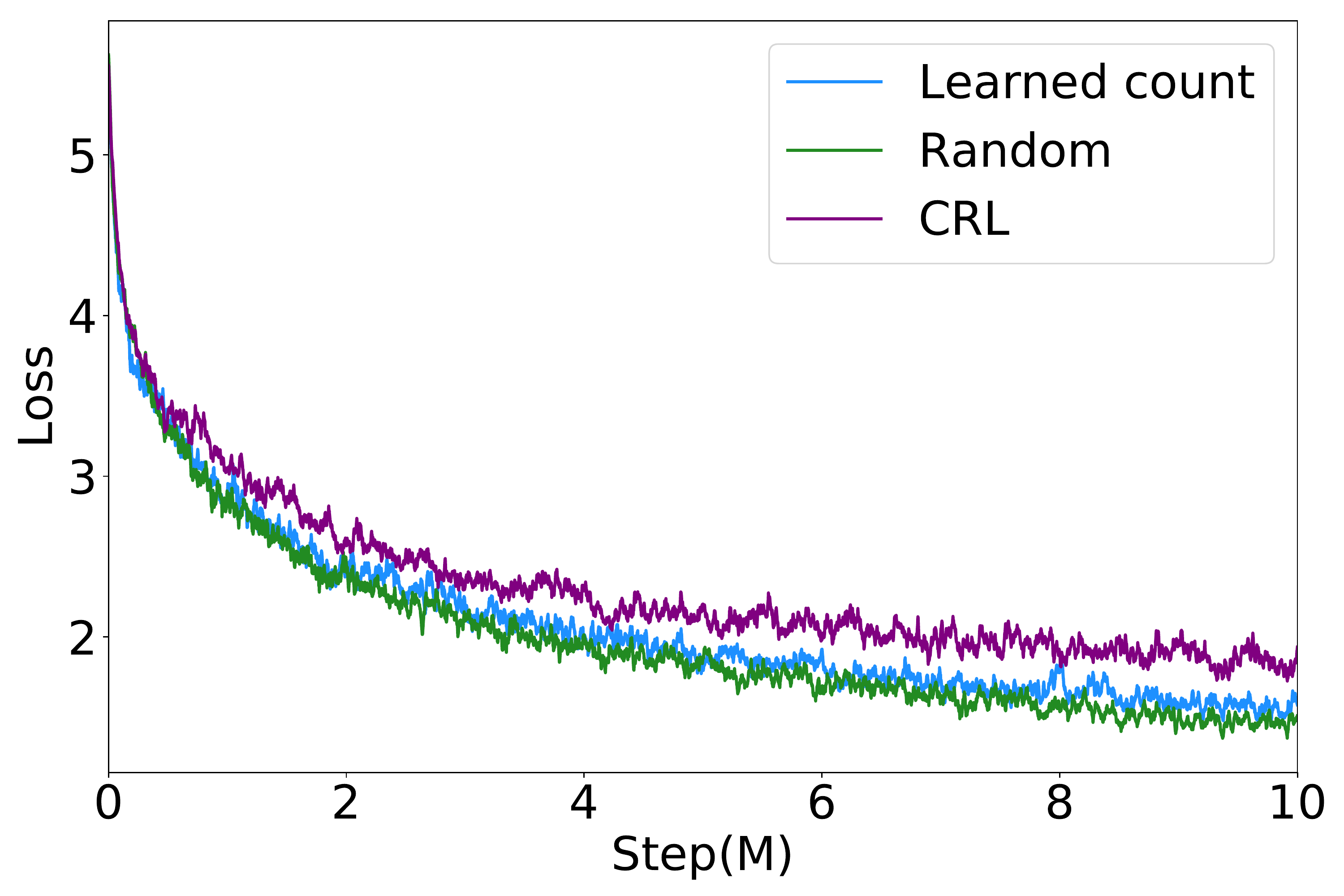}
\end{center}
\vspace{-20pt}
\caption{\small Plots of contrastive loss over time using different exploration methods. By treating the process of image gathering as an adversarial process, \model enables the procurement of diverse images, leading to larger contrastive loss.}
\label{fig:contrastive}
%\vspace{-5pt}
\end{figure}
\begin{figure}
\begin{center}
\includegraphics[width=\linewidth]{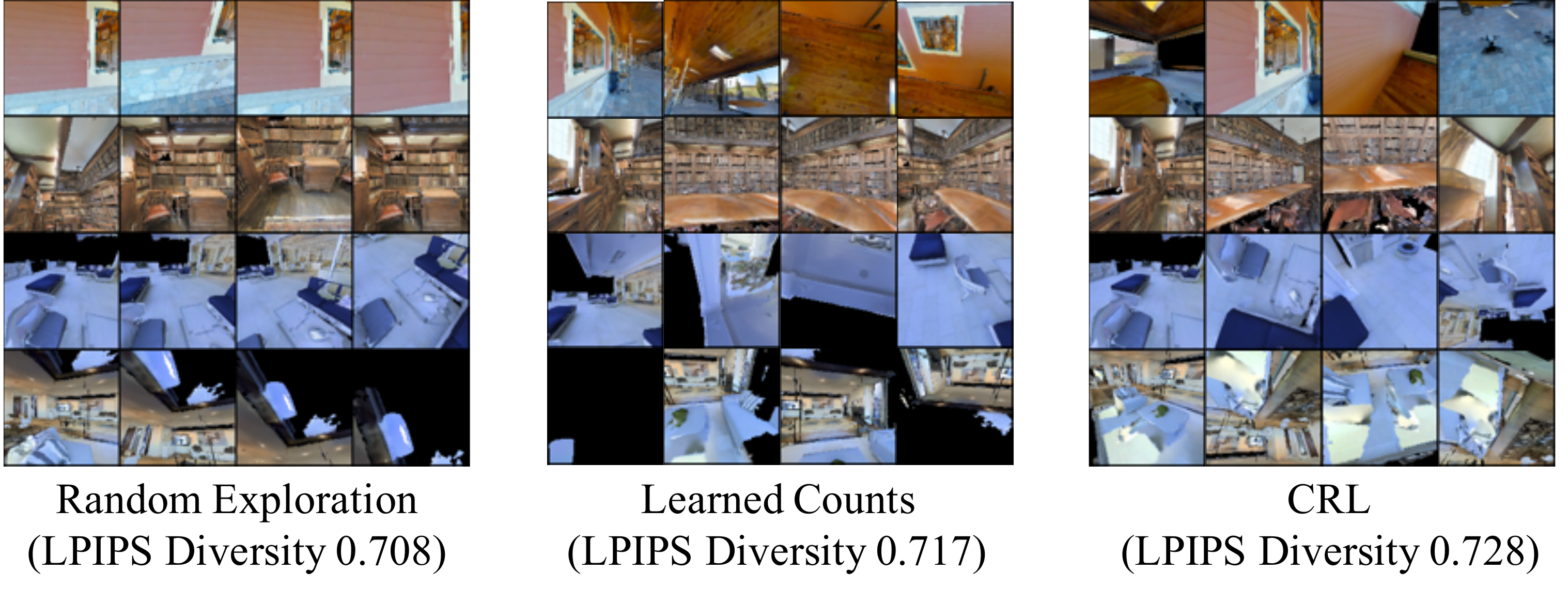}
\end{center}
\vspace{-15pt}
\caption{\small Illustration of data acquired for contrastive training utilizing either a random agent, a learned counts agent or \model on 4 different environments. Data collected by a random agent exhibit limited diversity, while data collected by a learned counts are diverse but not visually interesting (indicated by black backgrounds). Data collected by \model is diverse.}
\vspace{-15pt}
\label{fig:contrastive_data}
\end{figure}
\subsection{Experimental Setup}
\label{sect:embodied_rep}

To pretrain representations, we train different models on  the Habitat simulator using the Matterport3D dataset for 10 million interactions. In addition to \model, we consider the following set of baseline methods to obtain representations: 

\myparagraph{Exploration Strategies.} In \model, we rely on an intrinsically motivated policy to explore the surrounding world to train our contrastive model. We further compare using other approaches to collect data from the surrounding environment. We consider either using random actions to explore or using a learned counts-based exploration approach \citep{NIPS2017_3a20f62a}.

\myparagraph{Video Game Methods.} Concurrent to our work, recent work has explored learning state representation for reinforcement learning in static video game environments.  These works assume the presence of a statically collected dataset of experiences, and are customized to non-realistic video game settings. We compare with one such recent approach, augmented temporal constrast (ATC) \citep{stooke2020decoupling}, where we utilize the exploration policy of \model to explore the surrounding environment, and utilize ATC to learn a representation on top of the gathered images.

\myparagraph{Curiosity Objective.} Under \model, we may interpret other existing curiosity-based reinforcement learning approaches as representation learning objectives. We thus compare with one such objective, that of Random Network Distillation (RND) \citep{burda2018exploration}, which trains a model to regress the representations of a frozen network. We use RND to both incentive exploration, as well as learn a representation of the scene.

\myparagraph{Policy Based Representation.} An alternative approach to obtain representations is to utilize a reinforcement learning policy to learn representations of the environment. We thus compare with the representation learned by a PointNav policy trained on the Matterport training split. 

\myparagraph{ImageNet Pretraining.} We may also use existing large scale vision datasets to obtain our model. We thus provide a comparison study where we initialize our model using a ResNet50 pretrained on ImageNet.

All models are trained in Pytorch using PPO for 10M frames with the Adam optimizer. Hyperparameters for both representation pretraining and downstream evaluation are provided in the supplement.

\begin{table*}
\small\setlength{\tabcolsep}{5.5pt}
\centering
\caption{Full results of comparisons of embodied navigation with learned interactive representations average across 5 seperate seeds (with standard error in parentheses). Policies are evaluated on the test set of ImageNav and ObjectNav tasks and are trained for 10 million frames in each environment. We report the mean across 3 different seeds and report results of individual runs in the supplement. We consider either training an RL agent from scratch, utilizing existing representation learning methods (ATC \citep{stooke2020decoupling}, RND \citep{burda2018exploration} and contrastive learning) or utilizing supervised weights (PointNav Policy, ImageNet Initialization). RL agents initialized from pretrained weights have representations frozen, while all weights in the from scratch RL agent are trained.}
\vspace{-5pt}
\begin{tabular}{l|l|l|cccc}
    \toprule
    Environment & Category & Method & SPL$\uparrow$ & Soft SPL$\uparrow$ & Success$\uparrow$ & Goal Distance$\downarrow$ \\
     \midrule
    \multirow{7}{*}{ImageNav} & From Scratch & & 0.0207 (0.0012) & 0.173 (0.007) & 0.039 (0.003) & 4.85 (0.04)  \\
     \cmidrule{2-7}
     &  Other Representation & RND \citep{burda2018exploration} &  0.0158 (0.0027) & 0.124 (0.013) & 0.029 (0.004) & 5.29 (0.08)\\
     & Learning Algorithms & ATC \citep{stooke2020decoupling} & 0.0268 (0.0029) & 0.172 (0.013) & 0.059 (0.004) & 4.63 (0.04)\\
     \cmidrule{2-7}
     & \multirow{3}{*}{Contrastive Learning} & Random Exploration & 0.0285 (0.0014) & 0.195 (0.010) & 0.054 (0.003) & 4.68 (0.04) \\
     & & Learned Counts \citep{NIPS2017_3a20f62a}  & 0.0277 (0.0030) & 0.183 (0.011) & 0.057 (0.003) & \textbf{4.54 (0.08)} \\
     & & CRL (ours) & \textbf{0.0324 (0.0018)} & \textbf{0.219 (0.005)} & \textbf{0.058 (0.002)} & 4.55 (0.04) \\
    \cmidrule{2-7}
     & \multirow{2}{*}{Supervised} & PointNav Policy & 0.0254 (0.0021) & 0.187 (0.020) & 0.048 (0.002) & 4.66 (0.03) \\
     & & ImageNet Initialization & 0.0193 (0.0042) & 0.143 (0.022) & 0.050 (0.007) & 4.61 (0.01) \\
     \midrule
     \multirow{7}{*}{ObjectNav} & From Scratch & & 0.0010 (0.0006) & 0.037 (0.008) & 0.003 (0.002) & 7.94 (0.44)  \\
     \cmidrule{2-7}
     &  Other Representation & RND \citep{burda2018exploration} & 0.0000 (0.0000) & 0.007 (0.001) & 0.000 (0.000) & 7.96 (0.13) \\
     & Learning Algorithms & ATC \citep{stooke2020decoupling} & 0.0020 (0.0014) & 0.058 (0.013) & 0.003 (0.002) & 8.32 (0.27) \\
     \cmidrule{2-7}
     & \multirow{3}{*}{Contrastive Learning} & Random Exploration & 0.0042 (0.0007) & 0.076 (0.010) & 0.011 (0.002) & 7.39 (0.19) \\
     & & Learned Counts \citep{NIPS2017_3a20f62a}  & 0.0079 (0.0013) & 0.110 (0.008) & 0.026 (0.004) & 7.49 (0.17) \\
     & & CRL (ours) & \textbf{0.0144 (0.0046)} & \textbf{0.119 (0.007)} & \textbf{0.040 (0.019)} & 7.33 (0.13) \\
    \cmidrule{2-7}
     & \multirow{2}{*}{Supervised} & PointNav Policy &  0.0390 (0.0011) & 0.094 (0.005) & 0.007 (0.002) & \textbf{7.29 (0.08)}\\
     & & ImageNet Initialization & 0.0064 (0.0021) & 0.062 (0.004) & 0.010 (0.003) & 7.91 (0.10)\\
    \bottomrule
\end{tabular}
\label{tbl:main_tbl_rl}
\vspace{-10pt}
\end{table*}

\subsection{Visual Exploration}

\label{sect:visual_explore}

We first assess the ability of each method to actively explore the environment around it. We report the average number of $0.01 \times 0.01$ tiles explored (as measured by $x$, $y$ position in the simulator) in a given scene over the course of training in \fig{fig:num_tile}. We find that \model learns to explore well, outperforming both random policies and our curiosity baseline (RND) as well a PointNav policy trained explicitly to navigate around the environment. We find that \model performs similarily in terms of exploration to a learned counts-based exploration approach, but note that the learned counts-based exploration is explicitly encouraged to explore the surrounding environment, maintaining a count of explored tiles using a learned hash map, while \model encourages the policy to gather diverse data for representation learning.

We next evaluate the ability of \model to gather diverse data to train our contrastive model. We compare utilizing either random exploration, a learned counts-based method, or \model to obtain data to train a contrastive model. In \fig{fig:contrastive}, we plot contrastive loss curves obtained by utilizing data collected by each method. We find that since \model is trained to adversarialy generate data for the contrastive model, the overall contrastive loss is significantly higher in later stages of training. We further visualize batches of image data collected from different methods in \fig{fig:contrastive_data}, observing a high degree of visual diversity in the data collected through \model. We quantitatively observe larger diversity utilizing the LPIPS diversity metrics \citep{zhu2018multimodal} with details in \sect{sect:quant_diversity}. %Qualitatively, data collected through \model is more diverse than data collected from other approaches.

\begin{figure}
\begin{center}
\includegraphics[width=0.8\linewidth]{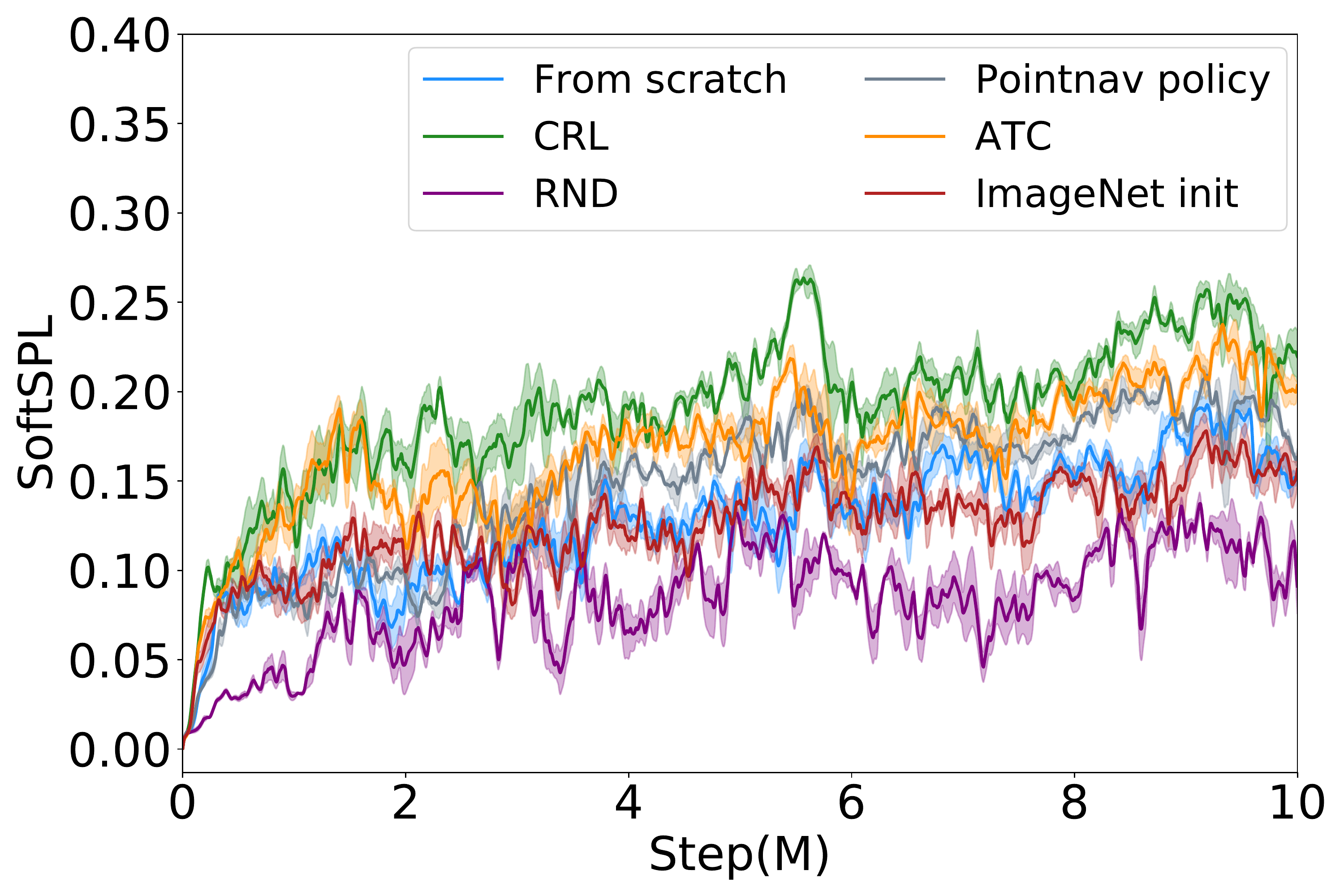}
\end{center}
\vspace{-15pt}
\caption{\small Plot of training SoftSPL over training steps of \model compared to other visual representation learning methods across 5 separate seeds in reinforcement learning. \model performs significantly better than initialization from scratch and outperforms all other methods on ImageNav in Gibson.}
\label{fig:imagenav_reward}
\vspace{-20pt}
\end{figure}
\begin{figure}
\begin{center}
\includegraphics[width=0.8\linewidth]{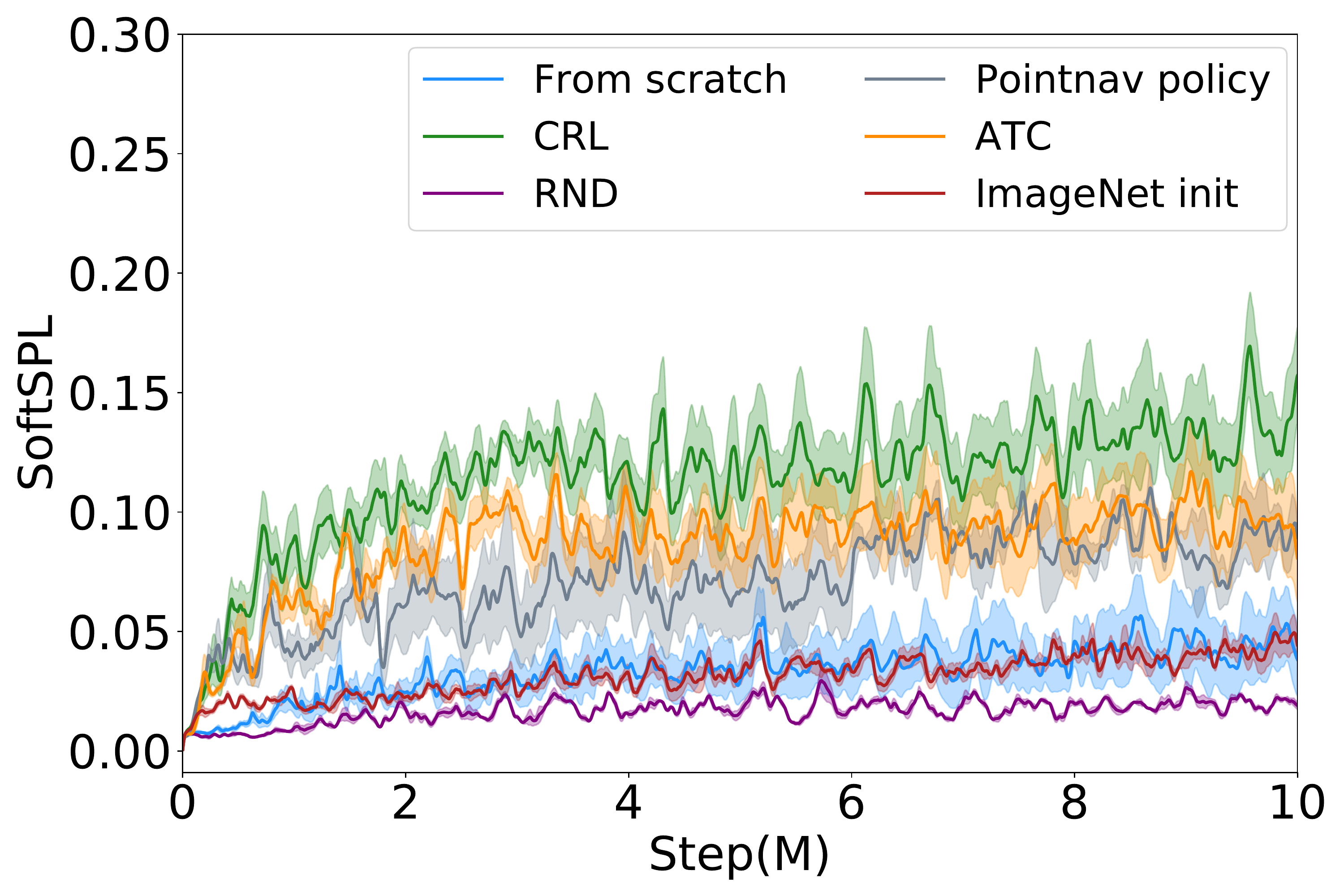}
\end{center}
\vspace{-15pt}
\caption{\small Plot of SoftSPL over training steps of \model compared to other visual representation learning methods across 5 separate seeds in reinforcement learning. \model performs significantly better than initialization from scratch and outperforms all other methods on ObjectNav in Matterport3D.}
\label{fig:objectnav_reward}
\vspace{-15pt}
\end{figure}
\begin{figure}
\begin{center}
 \includegraphics[width=0.8\linewidth]{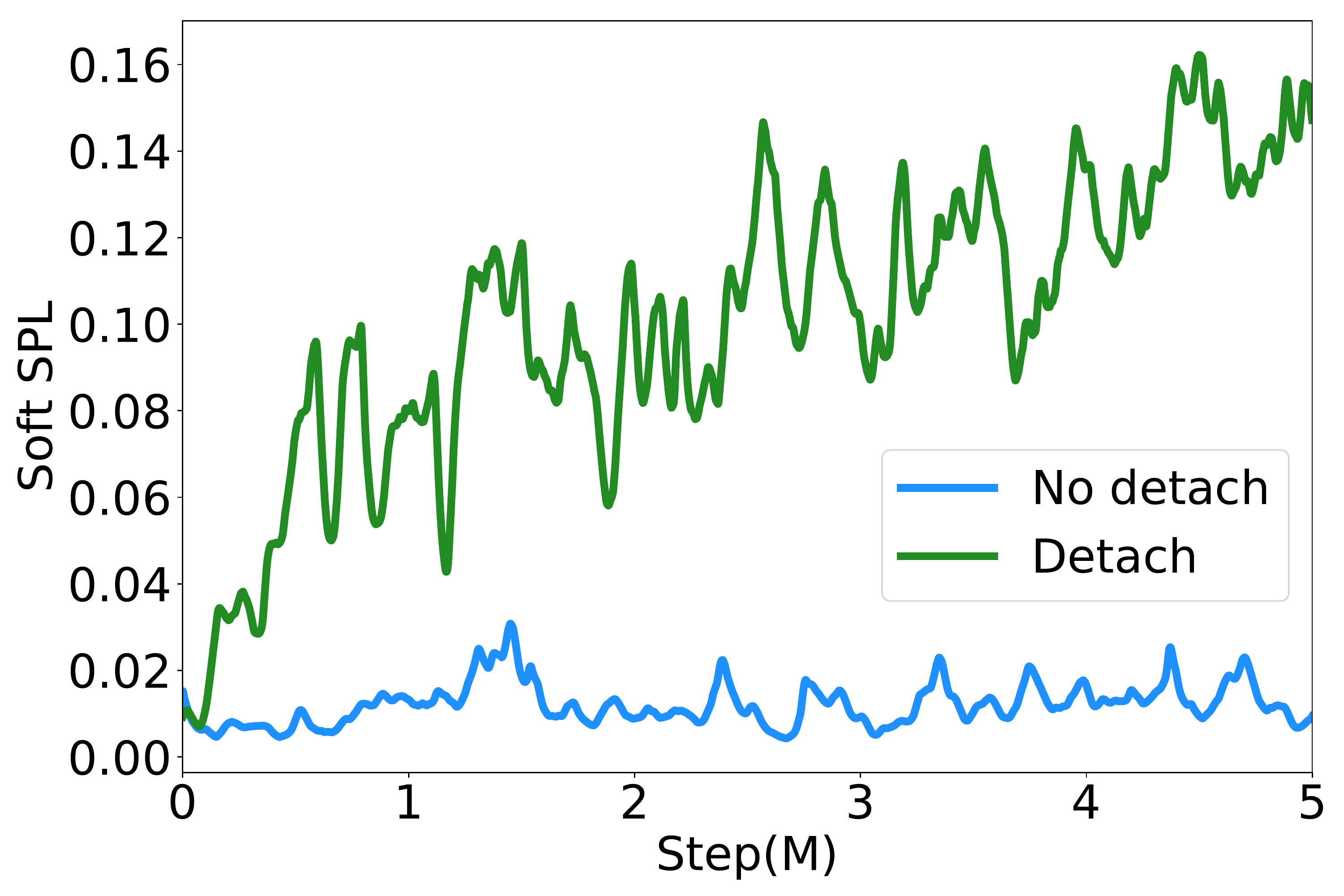}
\end{center}
\vspace{-15pt}
\caption{\small Plot of SoftSPL of a reinforcement learning agent on ObjectNav when the visual representation of \model is frozen or not frozen. Due to noisy gradient updates, when weights of \model are not frozen, performance deteriorates significantly.}
\label{fig:objectnav_detach}
\vspace{-20pt}
\end{figure}

\subsection{Semantic Navigation with RL}
\label{sect:rl}

Next, we investigate how each of the learned repreesentations \sect{sect:embodied_rep} can be utilized to learn effective RL policies for semantic navigation in the data-efficient setting. 

\myparagraph{Setup.} We measure reinforcement learning using the standard ImageNav task on the Gibson environment, and ObjectNav task on the Matterport3D environment included in Habitat \citep{habitat19arxiv}. Since we aim to validate the efficacy of learned visual representations, we train reinforcement learning policies using only 256$\times$256 RGB inputs, processed using learned representations, for 10 million frames. This setting is thus more challenging setting than typically evaluated in \citep{habitat19arxiv} as we assume the absence of either depth or robot localization information which is often given.

\myparagraph{Metrics.} We report standard metrics of visual navigation. We report tasks success, success weighted by path length (SPL) \citep{anderson2018evaluation}, soft SPL (success weighted by path length \citep{datta2020integrating}, but with a softer success criterion), and distance to goal. We utilize default criteria defined in \citep{habitat19arxiv}.

\myparagraph{Baselines.} We compare learning policies with representations from each approach described in \sect{sect:embodied_rep}.  We consider utilizing representations generated via different exploration strategies (Random, Learned Counts), using video game methods (ATC), changing the underlying curiosity objective (RND), extracted by an RL policy (PointNav), and weights initialized from ImageNet. In each setting, we freeze convolutional weights, which we find crucial for good performance.  We further compare with training an RL policy trained end-to-end entirely from scratch (From Scratch), including convolutional weights. We found that freezing weights for the From Scratch policy  significantly dropped performance (a fall of over 0.07 SoftSPL).

\myparagraph{Results.} We run each separate representation learning approach across 3 different random seeds, with mean performance across each metric reported in \tbl{tbl:main_tbl_rl}. Following \citep{anderson2018evaluation}, we recommend primarily looking at SPL and SoftSPL as metrics of performance. On both ImageNav and ObjectNav, we find that \model performs the best. Overall, we find that representations learned through contrastive learning lead to the best reinforcement learning performance. Subsequently, we find that weights from either PointNav weights or ATC perform better than training a policy from scratch. Surprisingly, we find that utilizing ImageNet does not appear to improve reinforcement learning performance significantly.

We visualize training SoftSPL across ImageNav in \fig{fig:imagenav_reward} and across ObjectNav in \fig{fig:objectnav_reward}. Similar to our reported metrics, we find that \model leads to the highest early increases in SoftSPL. Following that, we find that using weights from a trained PointNav policy or an ImageNet pretrained policy gives boosts to performance compared to a randomly initialized policy.

We note that although our result values are low, they are in line with those found in the 2020 Habitat navigation challenge\footnote{https://aihabitat.org/challenge/2020/}.  Furthermore, we note that we study a harder version of both semantic navigation tasks -- we only utilize RGB inputs to train our reinforcement learning policy, and do not assume information about either ground truth depth or the current localization of the policy.

\myparagraph{Importance of Detaching Gradients.} We further ablate the effect of freezing representations in \fig{fig:objectnav_detach}. We consider training a reinforcement learning policy on the object navigation task with or without freezing the weights of a reinforcement learning policy. We find that without freezing the weights of the convolutional network, SoftSPL increases significantly slower. 
\begin{figure}
\begin{center}
\includegraphics[width=0.8\linewidth]{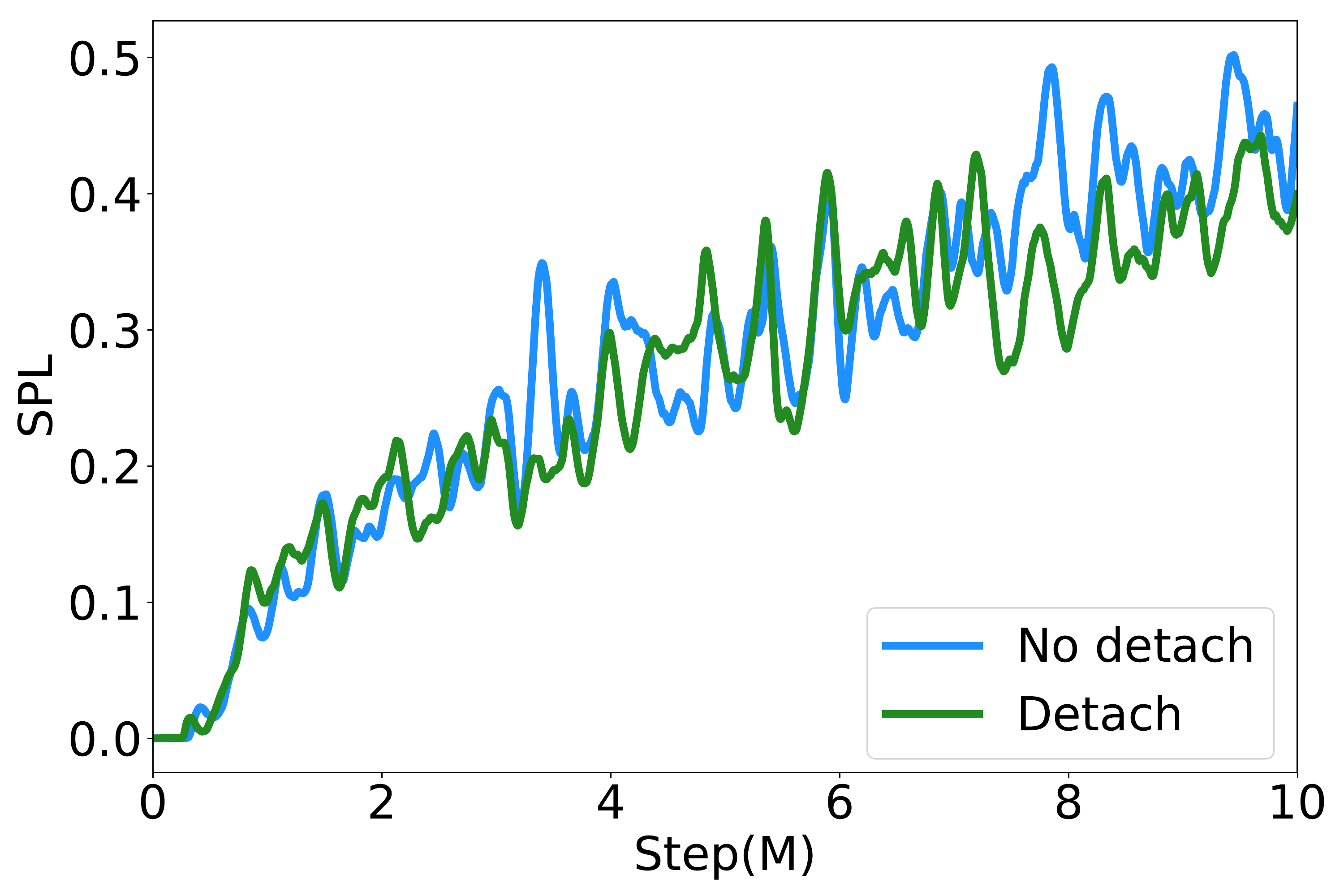}
\end{center}
\vspace{-20pt}
\caption{\small Plot of SPL of a reinforcement learning agent on PointNav, when convolution weights of a \emph{random} network are either frozen or not frozen. Surprisingly, we find that in PointNav, there is \textit{no} difference between each setting (we observe large differences in both ObjectNav and ImageNav tasks). Our results indicate that visual representations may not be effective enough in  the PointNav task to show performance gains in the limited data regime we study. }
\label{fig:pointnav_reward}
\vspace{-15pt}
\end{figure}

\myparagraph{What Doesn't Representation Learning Help On?} The most common evaluation task in Habitat \citep{habitat19arxiv} is the PointNav navigation task with a compass, where an embodied policy is instructed to navigate to a specific positional offset. Surprisingly, we find that learning a representation is not important in PointNav. In particular, in \fig{fig:pointnav_reward}, we initialize two separate policies from scratch and freeze the weights of the convolutional encoder of one policy. In both settings, we find that the overall PointNav SPL performance is \textit{identical}. We posit that in PointNav, in our data-efficient experimental setting, vision is not crucial to obtain good performance, since the policy is given a compass, but note that in the large-scale RL setting \citep{wijmans2019dd} shows that vision is indeed helpful for navigation.

\subsection{Instruction Navigation with Imitation Learning}
\label{sect:imitation}

\begin{table}
\small
\setlength{\tabcolsep}{5.5pt}
\centering
\caption{\small Comparison of performance of each pretrained representation on instruction following evaluated in unseen validation rooms.}
\vspace{-10pt}
\resizebox{\columnwidth}{!}{
\begin{tabular}{l|l|ccc}
    \toprule
    Setting & Method & SPL$\uparrow$ & Success$\uparrow$ & Goal Distance$\downarrow$ \\
     \midrule
     \multirow{5}{*}{Behavioral Cloning} & From Scratch & 0.138 & 0.152 & 9.17  \\
     & RND \citep{burda2018exploration} & 0.141 & 0.149 & 9.12 \\
     & ATC \citep{stooke2020decoupling} & 0.147 & 0.156 & 9.06\\
     & CRL (ours) & \textbf{0.157} & \textbf{0.169} & \textbf{8.77} \\
     \cmidrule{2-5}
     & Imagenet & 0.152 & 0.164 & 8.91 \\
     \midrule
     \multirow{5}{*}{Dagger} & From Scratch & 0.192  & 0.206  & 8.32 \\
     & RND \citep{burda2018exploration} & 0.187 & 0.200 & 8.23\\
     & ATC \citep{stooke2020decoupling} & 0.192 & 0.205 & 7.99 \\
     & CRL (ours) & \textbf{0.199} & \textbf{0.218} & \textbf{8.21} \\
     \cmidrule{2-5}
     & Imagenet  & 0.206 & 0.222 & 8.07 \\
    \midrule
    Random Agent & - & 0.0 & 0.0 & 10.23  \\
    \bottomrule
\end{tabular}
}
\label{tbl:main_tbl_bc}
\vspace{-10pt}
\end{table}
We next investigate how the different representation learning methods in \sect{sect:embodied_rep} can be utilized to aid  visual language navigation (VLN) via imitation learning.

\myparagraph{Setup.} We evaluate imitation learning using the vision language instruction benchmark introduced in \citep{krantz2020navgraph}. For simplicity, we utilize the base model and loss setting in \citep{krantz2020navgraph}, corresponding to training a Seq2Seq agent \citep{Sutskever2014Sequence} with or without Dagger \citep{dagger}. We utilize the author's implementation.   

\myparagraph{Metrics.} We use the same set of metrics defined in \sect{sect:rl}.  We report SPL, Success and goal distance  on the val-unseen split in \citep{krantz2020navgraph}, corresponding to unseen rooms, and report results on val-seen setting in the appendix.

\myparagraph{Baselines.} We compare representations learned from \model to those learned using either ATC or RND. We further compare representations from \model to utilizing weights from a supervised ImageNet model.

\myparagraph{Results.} We compare each learned representation when applied to imitation learning in \tbl{tbl:main_tbl_bc}. In both the behavioral cloning and Dagger settings, we find that utilizing \model obtains better performance than utilizing either random, RND, or ATC weights. We further find that \model obtains comparable performance to the Imagenet supervised model.

\subsection{Transfer to Real Image Recognition}

\label{sect:real_image}

Finally, we investigate to what extent our learned embodied representations, despite being learned entirely in simulation, can actually \textit{transfer} to real photographic scenes. 

\myparagraph{Setup.} To assess how representations transfer to realistic images, we utilize the Places dataset. We chose a subset of 59 class categories in Places corresponding to indoor room scenes (with selected class categories in the appendix). Following \citep{Zhang2016Colorful}, we then measure representations by fine-tuning a linear classifier on the final averaged-pooled features of our trained ResNet50 models. 

\begin{table}
\small
\setlength{\tabcolsep}{5.5pt}
\centering
\caption{\small Comparison of pretrained embodied representations in Habitat when transferred to the Places Dataset.}
\vspace{-10pt}
\begin{tabular}{l|cc|cc}
    \toprule
    Learning Objective & \multicolumn{4}{c}{Representation Accuracy} \\
    \midrule
       & \multicolumn{2}{c}{Policy Accuracy} & \multicolumn{2}{c}{Model Accuracy} \\
       & Top 1 & Top 5 & Top 1 & Top 5 \\
     \midrule
     Random Initialization & - & -  & 9.22 & 27.59 \\
     RND & 2.61 & 10.13 & 5.98  & 18.03 \\
     ATC & - & - & 14.83 & 40.61 \\

     CRL (ours) & \textbf{4.68}  & \textbf{18.32} &  \textbf{21.22} & \textbf{48.78}  \\
     \midrule
     PointNav & 4.31  & 15.27 & - & - \\
     ImageNet Pretraining & - & - & 54.59 & 85.15 \\
    \bottomrule
\end{tabular}
\label{tbl:habitat_policy_model}
\vspace{-10pt}
\end{table}
\begin{figure}
\begin{center}
\includegraphics[width=1.0\linewidth]{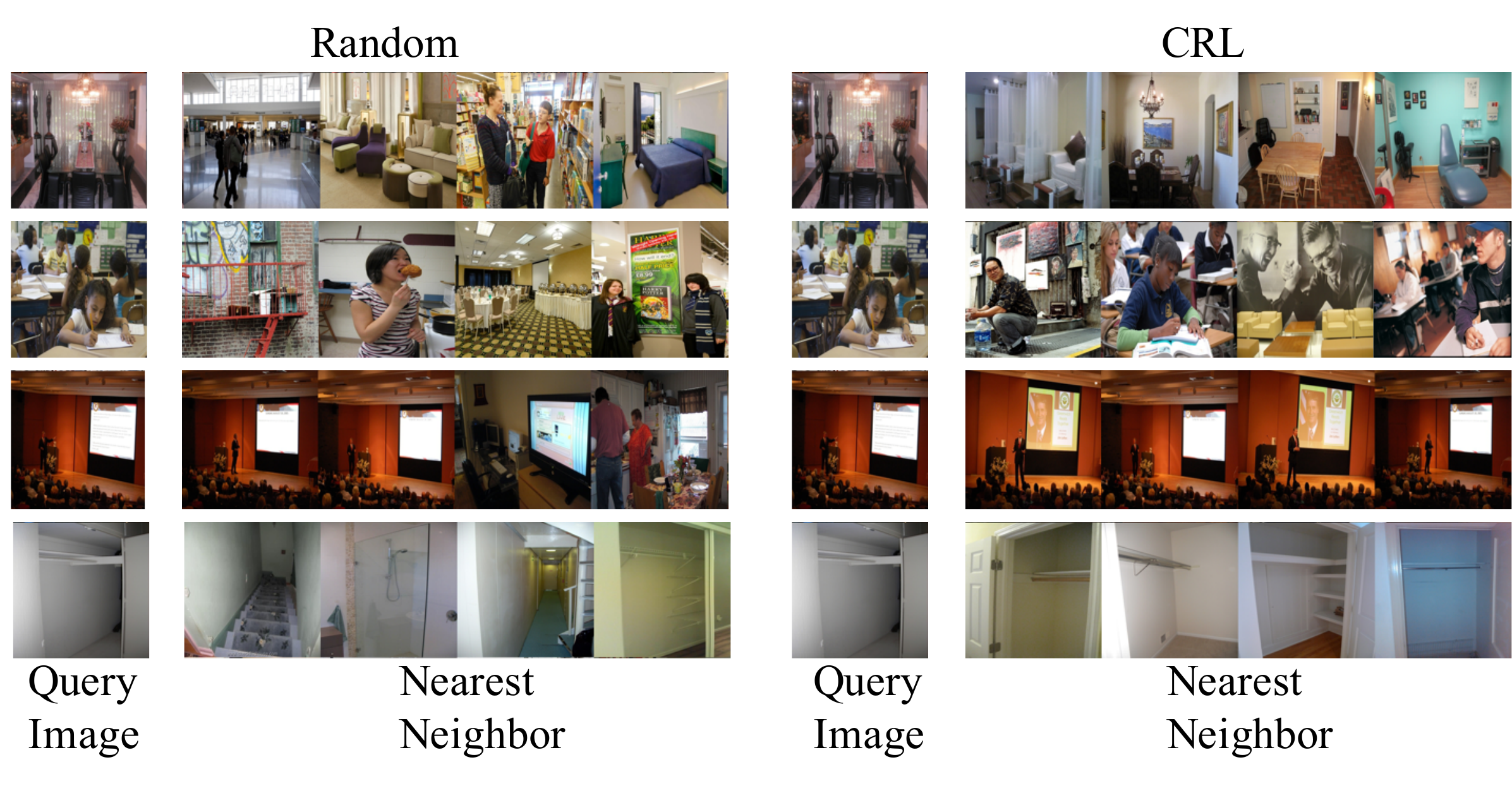}
\end{center}
\vspace{-20pt}
\caption{\small Comparative illustration of representation space nearest neighbors of \model  and a random network on room scenes in Places.}
\label{fig:nearest_neighbor_habitat}
\vspace{-17pt}
\end{figure}

\myparagraph{Baselines.} We compare with the same set of baselines as in \sect{sect:imitation}. For methods in which both an RL policy and model is learned, we evaluate the representations of both. To assess representation learning of RL policies, we also compare with the representations learned from a PointNav policy trained on the Habitat Matterport3D dataset.

\myparagraph{Results.} 
We report quantitative results from linear fine-tuning in \tbl{tbl:habitat_policy_model}. Overall, we find that \model learns representations that transfer best to real images, outperforming other approaches. Of our remaining methods, we observe that ATC learns the second best representation. We further find that image encoders of policies learn poor representations that do not transfer well to real images, with the visual encoder of the \model policy learning the best representation.

While our results are worse than those of an ImageNet supervised model, we emphasize that this is still strong performance on our task since our approach is trained \textit{entirely} in simulation without any \textit{supervision}. Qualitatively, we visualize representations from \model by finding the nearest neighbors, in learned representation space, of different images in the Places dataset in \fig{fig:nearest_neighbor_habitat}. Compared to a random network, we find more visually similar neighbors.

% We next evaluate on Habitat (\sect{sect:env}) to see if the visual representations learned through CRL transfer to real images from rooms in the Places dataset. We compare representations learned through CRL on RND, autoencoding, and colorization objectives as well as the PointNav objective.

% \label{sect:habitat_crl}
% \input{figText/nearest_neighbor.tex}
% \input{figText/habitat_table.tex}

% On real world images, similar to the VizDoom enviroment, we  find that CRL is able to enable good visual features in \tbl{tbl:habitat_table}. Trends between better model and visual representation hold, similar to in VizDoom with CRL, with the most effective representation learning algorithm being colorization. We find that CRL enables us to train representation learning models and policies to have \textbf{significantly better} visual features then agents optimizing the PointNav goal, with linear classification accuracy of 0.193 compared to 0.086 in PointNav for policies and 0.253 compared to 0.211 for the representation learning model. These linear classification accuracies are significantly better than random (0.084) and somewhat close to the linear classification accuracy of a colorization model directly trained on Places room scenes (0.324).

% To qualitatively study the visual representations in Habitat, we construct nearest neighbors in embedding space of a trained CRL colorization model and a random model in \fig{fig:nearest_neighbor}. We find that CRL trained models on Habitat are able cluster certain images in Places room scenes together such as beds.

\section{Conclusion}

In this paper, we proposed a generic framework to learn task-agnostic visual representations in embodied environments. Our learned representations enable promising transfer on downstream semantic and language guided navigation tasks, and further can transfer to visual recognition of real photos. We hope our proposed framework inspires future work towards learning both better task-agnostic representations and transferring to more complex embodied tasks~\cite{gan2021threedworld}.

\myparagraph{Acknowledgments.}
We thank MIT-IBM for support that led to this project. Yilun Du is funded by an NSF graduate research fellowship. We thank Dhruv Batra for giving helpful comments on the manuscript.
\bibliographystyle{ieee_fullname}
\bibliography{rep_reference,reference,egbib}

\newcommand{\appendixhead}%
{\centering\textbf{\Large Appendix: Curious Representation Learning for Embodied Intelligence}
\vspace{0.25in}}

\twocolumn[\appendixhead]
\maketitle

\appendix

We analyze our \model in the natural biological setting in \sect{sect:biological}. We provide hyper-parameter details for each setting reported in the paper in \sect{sect:training}, including pseudocode in \sect{sect:pseudo}. We then provide additional quantitative numbers of \sect{sect:quant}, quantifying diversity of gathered images in \sect{sect:quant_diversity}, individual reinforcement learning runs in \sect{sect:quant_rl}, and different imitation learning runs on the val-seen subset in \sect{sect:quant_imitation}.

\section{Biological Exploration}
\label{sect:biological}

Our interactive representation learning  approach in \sect{sect:visual_explore} still departs from real biological learning in several important ways. While real biological learning occurs in a single environment, in \sect{sect:visual_explore}, we assume several different environments run in parallel. Furthermore, while real biological exploration occurs contiguously and persistently across time, in \sect{sect:visual_explore}, we assume each environment has a maximum duration of exploration, after which the agent teleports to a new house environment.  While both assumptions are standard for RL training, in this section, we investigate how \model and other approaches in \sect{sect:visual_explore} behave in a realistic biological setting.

\begin{figure}
\begin{center}
\includegraphics[width=0.8\linewidth]{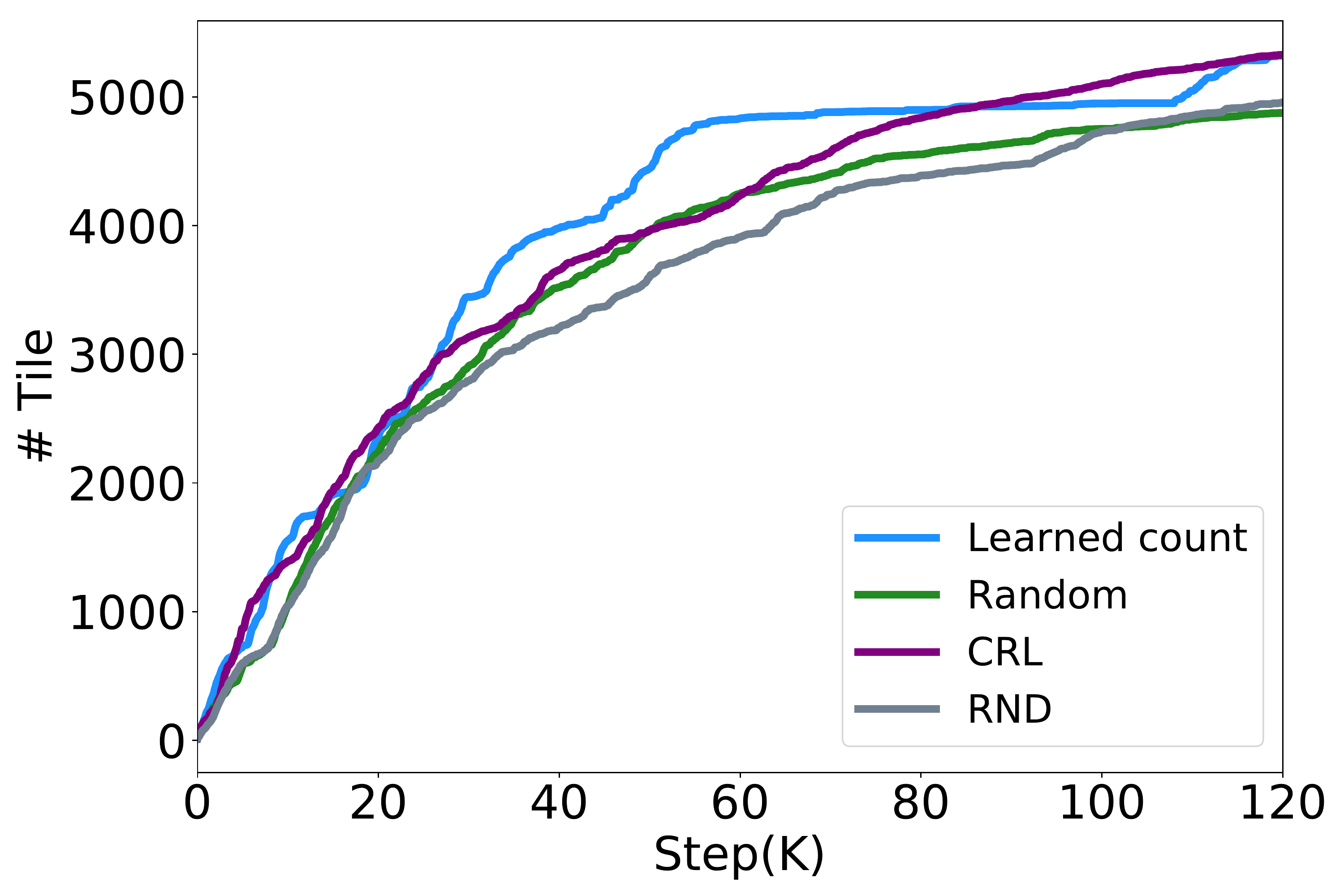}
\end{center}
\vspace{-20pt}
\caption{\small Plots of the average number of tiles explored in the biological setting where an agent is put in a single house environment. \model explores comparably to a learned counts-based method. }
\label{fig:num_tile_single}
\vspace{-10pt}
\end{figure}

\begin{figure}
\begin{center}
\includegraphics[width=0.8\linewidth]{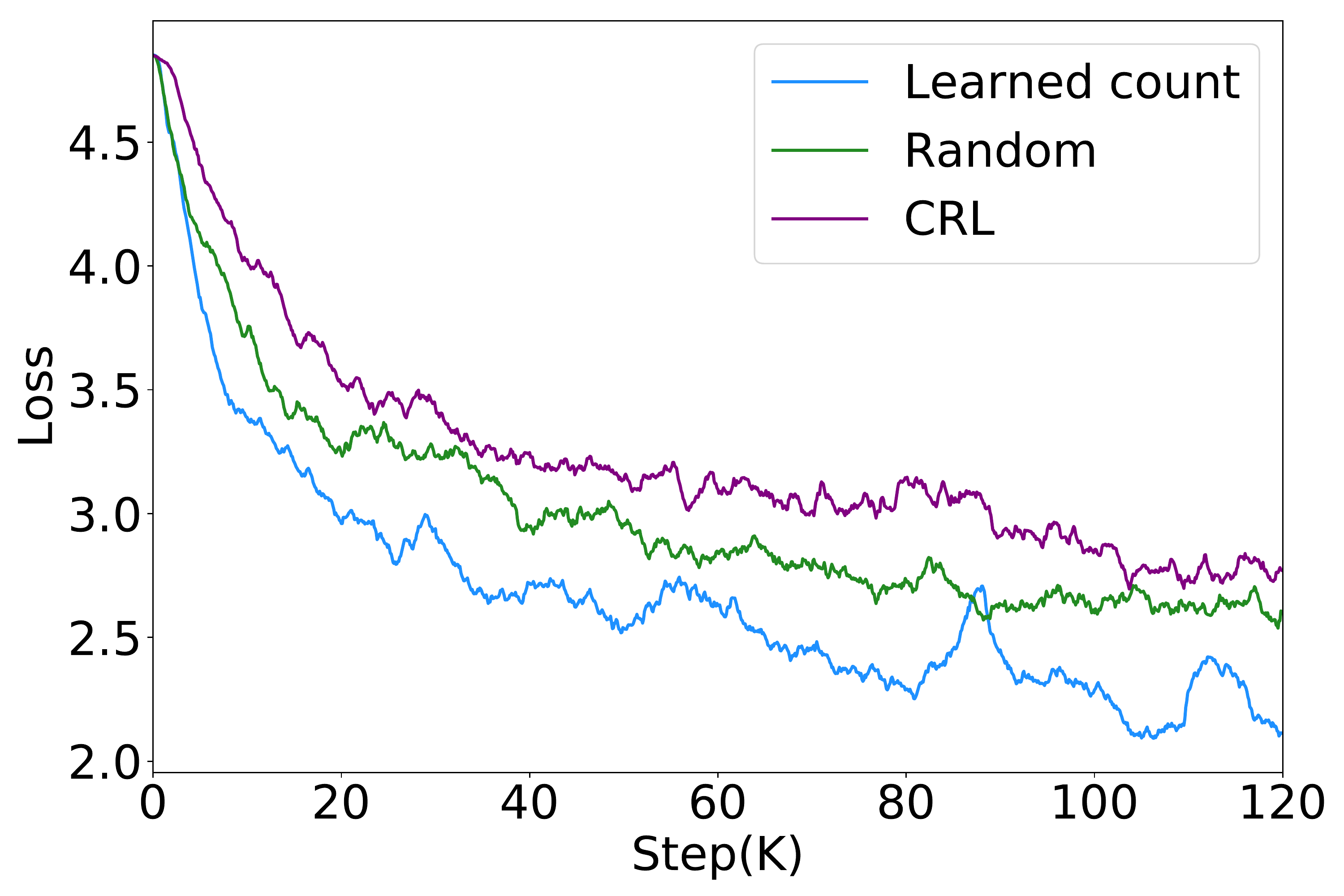}
\end{center}
\vspace{-20pt}
\caption{\small Plots of contrastive loss over time using different exploration methods in the biological learning setting of a single house environment. By treating the process of image gathering as an adversarial process, \model enables the procurement of diverse images, leading to larger contrastive loss, with larger differences than in \fig{fig:contrastive}.}
\label{fig:contrastive_single}
\vspace{-10pt}
\end{figure}

We train agents using a single house from the Habitat Matterport3D dataset, using a single environment process, and assume an infinitely long episode duration. We plot the number of tiles explored in that single environment in \fig{fig:num_tile_single} and and plot of contrastive loss on gathered images in \fig{fig:contrastive_single}. Overall, we find that \model gets a comparable number of tiles explored as the learned counts-based agent, but finds much more diverse images (indicated by a much higher contrastive loss in \fig{fig:contrastive_single}). When representations are evaluated for downstream real image recognition, we find the weights obtained from \model obtain a top 5 accuracy of 11.02 compared to 9.53 from random exploration and 8.98 from learned counts exploration.


\section{Training Details}
\label{sect:training}
We provide detailed hyper-parameters for each of the experiments in the paper. For RL  agents, we utilize the PPO implementation included with Habitat baselines \citep{habitat19arxiv}.

\subsection{Pretraining}

To pretrain representations, we utilize a total of 10M frames on the Habitat Matterport3D dataset. We train RL policies with 16 environments in parallel using PPO. Each individual episode has a maximum length of 500 steps. RL policies are trained with Adam with a learning rate of 0.0025, with generalized advantage estimation, an entropy coefficient of 0.01, discount factor 0.99, $\tau$ of 0.95, clip rate 0.2, with a data buffer size of 128 steps. Our RL policy is a LSTM network, with a single recurrent layer with hidden dimension of 512. Policies are updated for 4 epochs on images stored in the data buffer.

To train our representation learning model, we update the model at the same time as the RL policy, using the stored images in the data buffer. Models are tried with Adam with a learning rate of 0.0001. For contrastive learning models, we follow SimCLR \citep{chen2020simple} and utilize a temperature of 0.07, and the default ImageNet color augmentation, crop size, and horizontal flip augmentations. 

\subsection{ImageNav}

For the ImageNav task in the Habitat Gibson dataset, we train RL policies with 6 environments in parallel using PPO for 10 million frames. We utilize a ResNet50 to embed image goal observations (initialized with pretrained representations). Our RL policy is a recurrent network, with a single recurrent layer with hidden dimension of 512. Our RL policy is trained with PPO using Adam with a learning rate of 0.0025, a clip rate of 0.2, an entropy coefficient of 0.01, using generalized advantage estimation, with a discount factor of 0.99, $\tau$ of 0.95, and with a data buffer size of 64 steps. Policies are updated for 2 epochs on images stored in the data buffer.

\subsection{ObjectNav}

For the ObjectNav task in the Habitat Matterport3D dataset, we train RL policies with 16 environments in parallel using PPO for 10 million frames. We utilize a ResNet50 to embed image goal observations (initialized with pretrained representations). Our RL policy is a recurrent network, with a single recurrent layer with hidden dimension of size 512. Our RL policy is trained with PPO using Adam with a learning rate of 0.0025, a clip rate of 0.2, an entropy coefficient of 0.01, using generalized advantage estimation, with a discount factor of 0.99, $\tau$ of 0.95, and with a data buffer size of 64 steps. Policies are updated for 4 epochs on images stored in the data buffer.

\subsection{Language Imitation}

To train language imitation agents with either behavioral cloning or DAGGER, we directly utilize the authors' originally released repo, replacing convolutional encoders with or pretrained weights. We use Habitat-version 0.1.6 as our simulator for imitation learning.

\subsection{Places Images}

To finetune linear classifiers over ResNet50 average pooled features, we use the Adam optimizer with learning rate 0.001. We utilize early stopping to determine the number of training epochs to train linear classifiers, and train classifiers until the classification loss on the validation dataset increased (evaluated at the end of each training epoch).

We select the following classes in Places to apply classification over: abbey, alley, amphitheater, amusement\_park, aqueduct, arch, apartment\_building\_outdoor, badlands, bamboo\_forest, baseball\_field, basilica, bayou, boardwalk, boat\_deck, botanical\_garden, bridge, building\_facade, butte, campsite, canyon, castle, cemetery, chalet, coast, construction\_site, corn\_field, cottage\_garden, courthouse, courtyard, creek, crevasse, crosswalk, cathedral\_outdoor, church\_outdoor, dam, dock, driveway, desert\_sand, desert\_vegetation, doorway\_outdoor, excavation, fairway, fire\_escape, fire\_station, forest\_path, forest\_road, formal\_garden, fountain, field\_cultivated, field\_wild, garbage\_dump, gas\_station, golf\_course, harbor, herb\_garden, highway, hospital, hot\_spring, hotel\_outdoor, iceberg, igloo, islet, ice\_skating\_rink\_outdoor, inn\_outdoor, kasbah, lighthouse, mansion, marsh, mausoleum, medina, motel, mountain, mountain\_snowy, market\_outdoor, monastery\_outdoor, ocean, office\_building, orchard, pagoda, palace, parking\_lot, pasture, patio, pavilion, phone\_booth, picnic\_area, playground, plaza, pond, racecourse, raft, railroad\_track, rainforest, residential\_neighborhood, restaurant\_patio, rice\_paddy, river, rock\_arch, rope\_bridge, ruin, runway, sandbar, schoolhouse, sea\_cliff, shed, shopfront, ski\_resort, ski\_slope, sky, skyscraper, slum, snowfield, swamp, stadium\_baseball, stadium\_football, swimming\_pool\_outdoor, television\_studio, topiary\_garden, tower, train\_railway, tree\_farm, trench, temple\_east\_asia, temple\_south\_asia, track\_outdoor, underwater\_coral\_reef, valley, vegetable\_garden, veranda, viaduct, volcano, waiting\_room, water\_tower, watering\_hole, wheat\_field, wind\_farm, windmill, yard.

\subsection{Pseudocode}
\label{sect:pseudo}
For clarity, we present pseudocode describing the representation pretraining process of \model in \alg{alg:pseudocode}

\begin{algorithm}
\small
\begin{algorithmic}
    \STATE \textbf{Input:} Environment $E$, Representation learning model $M_\phi$, Policy $\pi_\theta$, Buffer $B$
    \STATE \emph{$\triangleright$ Train \model Model:}
    \WHILE{not converged}
    
    \STATE \emph{$\triangleright$ Gather information from the environment:}
    \FOR{sample $K$ steps}
    \STATE $\vx \leftarrow \text{get\_obs}(E)$
    \STATE $B = B \cup \vx$
    \STATE $a \leftarrow \pi_\theta(\vx)$
    \STATE $\text{step}(E, a)$ 
    \ENDFOR 
    \STATE \emph{$\triangleright$ Update $M_\phi, \pi_\theta$ using gathered data:}
    \STATE $\mathcal{L}_{\phi} = \mathcal{L}_{\text{Rep}}(M_\phi, B)$
    \STATE \emph{$\triangleright$ Compute loss for policy $\pi_\theta$ using reward equal to $\mathcal{L}_{\phi}$:}
    \STATE $\mathcal{L}_{\theta} = \mathcal{L}_{\text{PPO}}(\phi_\theta, B, \mathcal{L}_{\phi})$
    \STATE $\Delta \phi, \Delta \theta \gets \nabla_\phi \mathcal{L}_{\phi}, \nabla_\theta \mathcal{L}_{\theta}$
    \STATE \emph{Update $\phi, \theta$  using $\Delta \phi, \Delta \theta,$ through Adam:}
    \ENDWHILE
  \end{algorithmic}
 \caption{\model pretraining algorithm.}
 \label{alg:pseudocode}
 \end{algorithm}

\section{Additional Quantitative Results}
\label{sect:quant}

\subsection{Quantitative Measures of Exploration}
\label{sect:quant_diversity}
We quantitatively analyze the diversity of images found in the main paper Figure 5, utilizing the average distance between LPIPS embeddings of different images following \citep{zhu2018multimodal}. We collect 2048 across each exploration method, consisting of 128 seperate images gathered over 16 different environments. We find that using the exploration policy of \model obtains LPIPS diversity of 0.728 (0.001),  while the learned counts \citep{NIPS2017_3a20f62a} policy obtains LPIPS diversity of 0.717 (0.001) and random exploration obtains an LPIPS diversity of 0.708 (0.001), with standard error reported in parentheses  calculated across gathered trajectories. Quantitatively, \model leads to more diverse image gathering.

\subsection{Reinforcement Learning Quantitative Results}
\label{sect:quant_rl}

We provide a table of results across the first 3 evaluated seeds in ObjectNav and ImageNav in \tbl{tbl:main_tbl_rl_full_imagenav} and \tbl{tbl:main_tbl_rl_full_objectnav}. \model performs better than other approaches.

\subsection{Imitation Learning Quantitative Results}
\label{sect:quant_imitation}
We report imitation learning results on validation seen rooms in \tbl{tbl:main_tbl_bc_seen}. Using frozen representations from \model performs well.

\begin{table}
\small
\setlength{\tabcolsep}{5.5pt}
\centering
\caption{\small Comparison of performance of each pretrained representation on instruction following evaluated in seen validation rooms.}
\vspace{-10pt}
\resizebox{\columnwidth}{!}{
\begin{tabular}{l|l|ccc}
    \toprule
    Setting & Method & SPL$\uparrow$ & Success$\uparrow$ & Goal Distance$\downarrow$ \\
     \midrule
     \multirow{5}{*}{Behavioral Cloning} & From Scratch & 0.215 & 0.230  & 8.689   \\
     & RND \citep{burda2018exploration} & 0.210 &  0.228 & 8.536 \\
     & ATC \citep{stooke2020decoupling} &  0.210  &  0.223 & 8.379\\
     & CRL (ours) & \textbf{0.234} & \textbf{0.248}  & \textbf{8.364} \\
     \cmidrule{2-5}
     & Imagenet & 0.225 & 0.241  & 8.679  \\
     \midrule
     \multirow{5}{*}{Dagger} & From Scratch & 0.265  &  0.279 & 7.549 \\
     & RND \citep{burda2018exploration} & 0.284 & 0.302 & \textbf{7.044} \\
     & ATC \citep{stooke2020decoupling} & 0.273 & 0.290 & 7.117 \\
     & CRL (ours) & \textbf{0.295} & \textbf{0.316} & 7.441 \\
     \cmidrule{2-5}
     & Imagenet  & 0.267 & 0.281  & 7.399 \\
    \bottomrule
\end{tabular}
}
\label{tbl:main_tbl_bc_seen}
\vspace{-10pt}
\end{table}

\begin{table*}
\small\setlength{\tabcolsep}{5.5pt}
\footnotesize
\centering
\caption{Comparison of embodied navigation with learned interactive representations. Policies are evaluated on the test set of ImageNav tasks and are trained for 10 million frames in each environment. We report individual results of evaluated seeds. We consider either training an RL agent from scratch, utilizing existing representation learning methods (ATC \citep{stooke2020decoupling}, RND \citep{burda2018exploration} and contrastive learning) or utilizing supervised weights (PointNav Policy, ImageNet Initialization). RL agents initialized from pretrained weights have visual representations frozen, while all weights in the from scratch RL agent are trained.}
\vspace{-5pt}
\begin{tabular}{l|l|l|c|cccc}
    \toprule
    Environment & Category & Method & Seed & SPL$\uparrow$ & Soft SPL$\uparrow$ & Success$\uparrow$ & Goal Distance$\downarrow$ \\
     \midrule
     \multirow{40}{*}{ImageNav} &  \multirow{5}{*}{From Scratch} & \multirow{5}{*}{From Scratch} & 0  & 0.0238 & 0.191 & 0.036 & 4.98 \\
     & & & 1  &  0.0171 & 0.176 & 0.033 & 4.84 \\
      & & & 2  & 0.0237 & 0.185 & 0.051 & 4.76 \\
      & & & 3  & 0.0181 & 0.147 & 0.034 & 4.72 \\
      & & & 4  & 0.0206 & 0.166 & 0.038 & 4.94 \\
     \cmidrule{2-8}
     & \multirow{10}{*}{Other Representation} & \multirow{5}{*}{RND \citep{burda2018exploration}}  & 0 & 0.0166 & 0.119 & 0.044 & 5.09\\
     & \multirow{10}{*}{Learning Algorithms} & & 1 & 0.0050 & 0.082 & 0.020 & 5.32 \\
     & & & 2 & 0.0172 & 0.151 & 0.020 & 5.12 \\
     & & & 3 & 0.0167 & 0.101 & 0.024 & 5.62 \\
     & & & 4 & 0.0234 & 0.166 & 0.038 & 4.94 \\
     \cmidrule{3-8}
     & &  \multirow{5}{*}{ATC \citep{stooke2020decoupling}}  & 0 & 0.0183 & 0.133 & 0.060 & 4.59 \\
     & &   & 1 & 0.0339 & 0.209 & 0.063 & 4.69 \\
     & &   & 2 & 0.0231 & 0.180 & 0.043 & 4.64\\
     & &   & 3 & 0.0350 & 0.190 & 0.063 & 4.84\\
     & &   & 4 & 0.0237 & 0.146 & 0.070 & 4.49\\
     \cmidrule{2-8}
     & \multirow{15}{*}{Contrastive Learning} &   \multirow{5}{*}{Random} & 0 & 0.0320 & 0.204 & 0.058 & 4.70 \\
     & &   & 1 &  0.0315 & 0.198 & 0.054 & 4.73 \\
     &  &   & 2 & 0.0268 & 0.192 & 0.046 & 4.83 \\
     & &   & 3 &  0.0238 & 0.155 & 0.061 & 4.58 \\
     &  &   & 4 & 0.0277 & 0.217 & 0.048 & 4.60 \\
     \cmidrule{3-8}
     &  &   \multirow{5}{*}{Learned Counts \citep{NIPS2017_3a20f62a}} & 0 & 0.0320 & 0.193 & 0.053 & 4.75 \\
     & &   & 1 & 0.0210 & 0.178 & 0.057 & 4.30 \\
     &  &   & 2 & 0.0300 & 0.206 & 0.047 & 4.70 \\
     & &   & 3 & 0.0367 & 0.200 & 0.069 & 4.34 \\
     &  &   & 4 & 0.0192 & 0.138 & 0.056 & 4.61 \\
     \cmidrule{3-8}
     &  &   \multirow{5}{*}{CRL (ours)} & 0 & 0.0274 & 0.203 & 0.053 & 4.61 \\
     & &   & 1 & 0.0348 & 0.225 & 0.058 & 4.58  \\
     &  &   & 2 & 0.0313 & 0.239 & 0.051 & 4.53  \\
     &  &   & 3 & 0.0306 & 0.222 & 0.054 & 4.33  \\
     &  &   & 4 & 0.0364 & 0.227 & 0.064 & 4.59  \\
     \cmidrule{2-8}
     & \multirow{8}{*}{Supervised} & \multirow{3}{*}{PointNav Policy} & 0 & 0.0212 & 0.143 & 0.051 & 4.61\\
     & &   & 1 & 0.0249 & 0.192 & 0.048 & 4.63  \\
     &  &   & 2 &  0.0302 & 0.227 & 0.044 & 4.74\\
     \cmidrule{3-8}
     &  &   \multirow{5}{*}{ImageNet Initialization} & 0 & 0.0211 & 0.151 & 0.058 & 4.62  \\
     & &   & 1 & 0.0179 & 0.173 & 0.044 & 4.61  \\
     &  &   & 2 &  0.0315 & 0.175 & 0.066 & 4.63\\
     &  &   & 3 &  0.0229 & 0.172 & 0.061 & 4.56\\
     &  &   & 4 &  0.0020 & 0.044 & 0.021 & 4.61\\
    \bottomrule
\end{tabular}
\label{tbl:main_tbl_rl_full_imagenav}
\vspace{-10pt}
\end{table*}

\begin{table*}
\small\setlength{\tabcolsep}{5.5pt}
\footnotesize
\centering
\caption{Comparison of embodied navigation with learned interactive representations. Policies are evaluated on the test set of ObjectNav tasks and are trained for 10 million frames in each environment. We report individual results of the  evaluated seeds. We consider either training an RL agent from scratch, utilizing existing representation learning methods (ATC \citep{stooke2020decoupling}, RND \citep{burda2018exploration} and contrastive learning) or utilizing supervised weights (PointNav Policy, ImageNet Initialization). RL agents initialized from pretrained weights have visual representations frozen, while all weights in the from scratch RL agent are trained.}
\vspace{-5pt}
\begin{tabular}{l|l|l|c|cccc}
    \toprule
    Environment & Category & Method & Seed & SPL$\uparrow$ & Soft SPL$\uparrow$ & Success$\uparrow$ & Goal Distance$\downarrow$ \\
     \midrule
     \multirow{40}{*}{ObjectNav} &  \multirow{5}{*}{From Scratch} & \multirow{5}{*}{From Scratch} & 0 & 0.0000 & 0.0475 & 0.000 & 8.06  \\
     & & & 1  & 0.0032 & 0.0487 & 0.010 & 6.93  \\
      & & & 2  & 0.0000 & 0.0049 & 0.000 & 7.69 \\
      & & & 3  & 0.0016 & 0.0533 & 0.003 & 7.28 \\
      & & & 4  & 0.0000 & 0.0292 & 0.000 & 9.73 \\
     \cmidrule{2-8}
     & \multirow{10}{*}{Other Representation} & \multirow{5}{*}{RND \citep{burda2018exploration}}  & 0 & 0.0000 & 0.0043 & 0.000 & 8.06\\
     & \multirow{10}{*}{Learning Algorithms} & & 1 & 0.0000 & 0.0070 & 0.000 & 7.76 \\
     & & & 2 & 0.0000 & 0.0046 & 0.000 & 7.33\\
     & & & 3 & 0.0000 & 0.0533 & 0.003 & 7.28\\
     & & & 4 & 0.0000 & 0.0081 & 0.000 & 7.53\\
     \cmidrule{3-8}
     & &  \multirow{5}{*}{ATC \citep{stooke2020decoupling}}  & 0 & 0.0080 & 0.0881 & 0.010 & 8.37 \\
     & &   & 1 & 0.0000 & 0.0923 & 0.000 & 7.93\\
     & &   & 2 & 0.0000 & 0.0525 & 0.000 & 8.23\\
     & &   & 3 & 0.0020 & 0.0232 & 0.003 & 7.64\\
     & &   & 4 & 0.0000 & 0.0334 & 0.000 & 9.44\\
     \cmidrule{2-8}
     & \multirow{15}{*}{Contrastive Learning} &   \multirow{5}{*}{Random} & 0 &  0.0041 & 0.0888 & 0.010 & 7.76 \\
     & &   & 1 & 0.0034 & 0.0963 & 0.010 & 7.35 \\
     &  &   & 2 &  0.0017 & 0.0380 & 0.003 & 6.59\\
     & &   & 3 & 0.0060 & 0.0669 & 0.010 & 7.52 \\
     &  &   & 4 &  0.0060 & 0.0904 & 0.020 & 7.73\\
     \cmidrule{3-8}
     &  &   \multirow{5}{*}{Learned Counts \citep{NIPS2017_3a20f62a}} & 0 & 0.0068 & 0.1174 & 0.029 & 7.26  \\
     & &   & 1 & 0.0075 & 0.1294 & 0.021 & 6.95 \\
     &  &   & 2 &  0.0069 & 0.0830 & 0.030 & 8.03 \\
     & &   & 3 & 0.0136 & 0.1244 & 0.040 & 7.47 \\
     &  &   & 4 & 0.0048 & 0.0972 & 0.010 & 7.75 \\
     \cmidrule{3-8}
     &  &   \multirow{5}{*}{CRL (ours)} & 0 & 0.0304 & 0.1326 & 0.120 & 6.97  \\
     & &   & 1 &  0.0084 & 0.1030 & 0.010 & 7.38 \\
     &  &   & 2 & 0.0128 & 0.1300 & 0.030 & 7.50 \\
     & &   & 3 &  0.0099 & 0.0960 & 0.000 & 7.74 \\
     &  &   & 4 &  0.0202 & 0.1306 & 0.040 & 7.05\\
     \cmidrule{2-8}
     & \multirow{10}{*}{Supervised} & \multirow{3}{*}{PointNav Policy} & 0 & 0.0021 & 0.0872 & 0.003 & 7.47 \\
     & &   & 1 & 0.0064 & 0.0881 & 0.009 & 7.15  \\
     &  &   & 2 &  0.0310 & 0.1070 & 0.010 & 7.27\\
     \cmidrule{3-8}
     &  &   \multirow{5}{*}{ImageNet Initialization} & 0 &  0.0144 & 0.0610 & 0.020 & 8.14\\
     & &   & 1 &  0.0046 & 0.0730 & 0.010 & 7.82\\
     &  &   & 2 &  0.0081 & 0.0576 & 0.009 & 7.74 \\
     & &   & 3 &  0.0000 & 0.0503 & 0.000 & 7.66\\
     &  &   & 4 &  0.0047 & 0.0690 & 0.010 & 8.18 \\
    \bottomrule
\end{tabular}
\label{tbl:main_tbl_rl_full_objectnav}
\vspace{-10pt}
\end{table*}

% \appendix
% \renewcommand{\thesection}{A.\arabic{section}}
% \renewcommand{\thefigure}{A\arabic{figure}}
% \setcounter{section}{0}
% \setcounter{figure}{0}
% % Text to make the supplement
% % \input{text/appendix.tex}

% \bibliographystyle{ieee_fullname}
% \bibliography{rep_reference,reference,egbib}

\end{document}